\pdfoutput=1
\documentclass[11pt]{article}

\usepackage[final]{acl}

\usepackage{times}
\usepackage{latexsym}
\usepackage{algorithm}
\usepackage{algpseudocode}
\usepackage[T1]{fontenc}
\usepackage[utf8]{inputenc}

\usepackage{microtype}

\usepackage{inconsolata}

\usepackage{wrapfig}
\usepackage{graphicx}
\usepackage{amssymb}
\usepackage{amsmath}
\usepackage{subfigure}
\usepackage{booktabs}
\usepackage{multirow}
\usepackage{adjustbox}
\usepackage{colortbl}
\usepackage{subcaption}
\usepackage{placeins}
\usepackage{float}
\usepackage{dblfloatfix}
\title{\textsc{UNComp}: Can Matrix Entropy Uncover Sparsity? — A Compressor Design from an Uncertainty-Aware Perspective}

\author{
  \textbf{Jing Xiong}$^{1}$ \quad \textbf{Jianghan Shen}$^{2}$ \quad \textbf{Fanghua Ye}$^{3}$\thanks{Corresponding author: fanghua.ye.21@gmail.com} \quad \textbf{Chaofan Tao}$^{1}$ \\
  \textbf{Zhongwei Wan}$^{4}$ \quad \textbf{Jianqiao Lu}$^{1}$ \quad \textbf{Xun Wu}$^{5}$ \quad \textbf{Chuanyang Zheng}$^{6}$ \\
  \textbf{Zhijiang Guo}$^{8}$ \quad \textbf{Min Yang}$^{7}$ \quad \textbf{Lingpeng Kong}$^{1}$ \quad \textbf{Ngai Wong}$^{1}$ \\
  \footnotesize
  $^{1}$ The University of Hong Kong \quad 
  $^{2}$ Nanjing University \quad 
  $^{3}$ University College London \quad 
  $^{4}$ The Ohio State University \\ 
  \footnotesize
  $^{5}$ Microsoft Research Asia \quad 
  $^{6}$ The Chinese University of Hong Kong \quad 
  $^{7}$ SIAT, Chinese Academy of Sciences \\ 
  \footnotesize
  $^{8}$ The Hong Kong University of Science and Technology (Guangzhou) 
  \\
 \small{
   \textbf{Contact:} \href{junexiong@connect.hku.hk}{junexiong@connect.hku.hk}
 }
}

\newtheorem{lemma}{Lemma}

\begin{document}
\maketitle
\begin{abstract}

% Deploying Large Language Models (LLMs) for long-context inference is challenging due to high memory and computational demands. Existing Key-Value (KV) cache compression methods, which use heuristic sparsity patterns, suffer from two key limitations: 1) uniform compression across heads and layers degrades retrieval performance, and 2) inefficient prefill stages lack a unified criterion for compressing hidden states to KV caches. We propose \textsc{UNComp}, a two-stage framework leveraging matrix entropy to adaptively optimize compression. During prefill, \textsc{UNComp} dynamically compresses hidden states at the layer level, while in decoding, it applies adaptive compression ratios to KV cache heads based on identified sparsity patterns.  \textsc{UNComp} reduces KV cache size to 4.74\% of the original size, achieves a 60\% prefill speedup, and improves throughput by 6.4×, achieving a 45\(\times\) speedup in retrieval head identification. with virtually no performance loss. Notably, it outperforms full-size caches in needle-in-a-haystack tasks when compressed to 9.38\% of the original size. \textsc{UNComp} integrates seamlessly with existing KV compression methods and Flash Attention, offering a plug-and-play solution. Our codes are submitted with the paper and will be publicly available.

Deploying large language models (LLMs) for long-context inference remains challenging due to their substantial memory and computational demands. While techniques such as Key-Value (KV) cache compression are designed to reduce memory usage, they often neglect the structured sparsity inherent in the relationship between hidden states and their corresponding KV cache. In this work, we explore the role of uncertainty as a potential indicator of sparsity within LLMs. We propose \textsc{UNComp}, an uncertainty-aware framework that leverages \textit{truncated matrix entropy} to identify areas of low information content, thereby revealing sparsity patterns that can be used for adaptive compression. Unlike traditional methods that apply uniform compression, \textsc{UNComp} dynamically adjusts its approach to compression, guided by uncertainty measures that reflect the importance of various model components. Our analysis shows that sparsity patterns, when derived from uncertainty estimates, can be exploited to reveal special long-range dependencies, such as retrieval heads and retrieval layers. This perspective not only enhances our understanding of how compression can be optimized but also provides new insights into the inherent sparsity of LLMs during long-context inference. By focusing on uncertainty to analyze the sparsity pattern in detail, \textsc{UNComp} reduces the KV cache size to 4.74\% of the original, achieves a 6\% prefill speedup, and improves throughput by 6.4× — not only delivering strong lossless compression performance, but also validating the effectiveness of the underlying theoretical tool. We
release the code at \url{https://github.com/menik1126/UNComp.}

\end{abstract}

\section{Introduction}

The development of large language models (LLMs) drives remarkable progress in natural language processing~\citep{achiam2023gpt, kaplan2020scaling}, enabling tasks ranging from text generation to complex reasoning. However, deploying LLMs for long-context inference remains resource-intensive, as they demand substantial memory and computation~\citep{shazeer2017outrageously}. A commonly adopted optimization, the \textit{KV cache} \citep{pope2023efficiently, liu2024minicache}, stores keys and values from previous tokens to avoid redundant computations. Nonetheless, maintaining a full KV cache for long sequences imposes considerable memory overhead, which poses a bottleneck~\citep{liu2024minicache}.

To address this, several compression strategies emerge:
\textit{i}) \textbf{Eviction} \citep{ge2023model, zhang2024h2o, li2024snapkv, zhang2024pyramidKV},
\textit{ii}) \textbf{Merging} \citep{liu2024minicache, wan2024d2o, wang2024model, zhangcam},
\textit{iii}) \textbf{Quantization} \citep{hooper2024KVquant, zhang2024KV, liu2024kivi}, and
\textit{iv}) \textbf{Head Pruning} \citep{ainslie2023gqa, shazeer2019fast, liu2024deepseek, yu2024effectively, brandon2024reducing}. However, these methods typically operate on sparsity along a single axis—such as pruning attention heads or compressing individual layers—without fully capturing how sparsity emerges across the model’s hierarchical structure, overlooking its complexity from different layers and attention heads. To bridge this gap, we introduce Matrix Information Theory~\citep{zhang2023matrix} and propose \textit{truncated matrix entropy} as a unified framework that connects uncertainty and sparsity in a principled manner.

Specifically, existing methods~\citep{zhang2024h2o, xiao2023efficient} often overlook the sparsity pattern shared between hidden states—typically the outputs of the multi-layer perceptron (MLP)—and the KV cache. Rather than viewing sparsity as a localized artifact of attention mechanisms, we interpret it as an indicator of low-information regions distributed throughout the hidden states and KV cache. This shift—from the empirical observation layer and head sparsity~\citep{jiang2024minference, wu2024retrieval, xiao2024duoattention,cai2024pyramidkv} to uncertainty-aware compression—reveals latent redundancies during long-context inference and guides adaptive hidden state compression for faster prefill and KV cache eviction reducing.  \vspace{1mm} \noindent Our key contributions are as follows: \begin{enumerate} 

\item We propose a novel uncertainty-aware compression framework grounded in \textit{truncated matrix entropy}, uncovering the relationship between information compression patterns and sparsity patterns in LLMs. Furthermore, we provide a detailed empirical study of the connection between information compression patterns and compression algorithms.

% \vspace{-1mm}
\item We design a two-stage compression framework that jointly compresses hidden states and the KV cache. We are the first to indirectly compress the KV cache and accelerate the prefill stage by compressing the hidden states, achieving a 60\% speedup in the prefill stage, a 4.74\% compression ratio, and a 6.4$\times$ improvement in throughput with negligible performance degradation. 
% \vspace{-1mm}
\item We demonstrate that our method maintains performance even under aggressive compression regimes, including settings where heads are entirely removed. In the needle-in-a-haystack task, \textsc{UNComp} surpasses the full-size KV cache baseline at a 9.38\% compression ratio. \end{enumerate}

\section{Related Work}

\subsection{Attention-Based Token Eviction}
\label{Eviction Policy}
Early works identify distinctive attention patterns in the KV cache, such as the \textit{attention sink}~\citep{liu2024lost, xiao2023efficient}. In addition, prior studies~\citep{cai2024pyramidkv, yang2024pyramidinfer} empirically observe that sparsity increases with model depth. Distinct sparsity patterns—such as retrieval heads~\citep{xiao2024duoattention, wu2024retrieval}—are also observed across attention heads. However, the underlying mechanisms driving these sparsity patterns remain to be fully understood.

 Recent methods employ cumulative attention scores for selecting subsets of tokens~\citep{zhang2024pyramidKV, li2023compressing, jiang2024minference, zhang2024h2o, ge2023model, sheng2023flexgen, liu2024scissorhands, li2024snapkv}. These methods apply a fixed compression ratio to sparsity patterns observed in the layers of the KV cache. However, they overlook the compression of hidden states, as well as differences in retrieval behaviors across layers. This not only degrades the performance of \textit{retrieval layers}, but also misses the opportunity to accelerate computation in the prefill stage by failing to compress the hidden states.
% \vspace{-1mm}
\subsection{Information Compression Behavior}
The sparsity patterns discussed in Section~\ref{Eviction Policy} are related to the model's internal information compression behavior~\citep{feng2022rank}. In this paper, we provide a detailed categorization and analysis of the compression patterns across different parameter matrices. Recent work~\citep{deletang2023language} reveals that models exhibit spontaneous compression behavior during training. Similar phenomena are reported in~\citet{tao2024scaling, huang2024compression}. These observations, however, rarely connect information compression patterns with sparsity patterns. This motivates us to explore the model's information compression behavior—through entropy increases or decreases—as a means to analyze its sparsity patterns. To achieve this, we introduce Matrix Information Theory~\citep{zhang2023matrix}, a theoretical tool for understanding the information compression behavior.
% \vspace{-1mm}
\section{Method}
% \vspace{-1mm}
In this section, to explore the information compression patterns in the model, we derive the definition of \textit{truncated matrix entropy} and reveal the connection between information compression patterns and sparsity patterns, leading to compression strategy.
% \vspace{-1mm}
\subsection{Preliminary}
To define \textit{truncated matrix entropy}, we first introduce the matrix entropy of the token matrix (either from a layer or a head). Let the token matrix be \(\mathbf{X} = [\mathbf{x}_1, \mathbf{x}_2, \dots, \mathbf{x}_N]\), where \(\mathbf{x}_i \in \mathbb{R}^D\) is the \(i\)-th token vector, and \(N\) is the sequence length. The covariance matrix \(\mathbf{\Sigma}_{\mathbf{X}} \in \mathbb{R}^{D \times D}\) is computed as:
\begin{equation}
\small
\mathbf{\Sigma}_{\mathbf{X}} = \frac{1}{N-1} \sum_{i=1}^{N} (\mathbf{x}_i - \bar{\mathbf{x}})(\mathbf{x}_i - \bar{\mathbf{x}})^T,
\end{equation}
where \(\bar{\mathbf{x}}\) is the mean vector of the sequence \(\mathbf{X}\). Based on this covariance matrix, we derive the definition of matrix entropy. Specifically, following \citet{giraldo2014measures}, the matrix entropy based on  \(\mathbf{\Sigma}_{\mathbf{X}}\) is defined as:
\begin{lemma}
\textit{As \(\alpha \to 1\) (the entropy measure depending on \(\alpha\)), we obtain the definition of the von Neumann (matrix) entropy~\citep{von2013mathematische}}:
\label{eq:Lemma 3.1.}
\begin{equation}
\small
\label{eq:matrix_entropy}
H(\mathbf{\Sigma}_{\mathbf{X}}) = - \text{Tr} \left(  \mathbf{\Sigma}_{\mathbf{X}} \log \left(  \mathbf{\Sigma}_{\mathbf{X}} \right) \right).
\end{equation}
\end{lemma}

\begin{lemma}
\textit{Let} \(\mathbf{\Sigma}_{\mathbf{X}}\) \textit{be a symmetric positive definite matrix with eigenvalues} \(\sigma=(\sigma_1,\sigma_2, \dots, \sigma_D)^T\). \textit{The matrix entropy of} \( \mathbf{\Sigma}_{\mathbf{X}} \) \textit{can be expressed as}:
\label{eq:Lemma 3.2.}
%\vspace{-2mm}
\begin{equation}
\small
H(\mathbf{\Sigma}_{\mathbf{X}}) = - \sum_{i=1}^{D} \sigma_i \log \sigma_i,
\end{equation}
\end{lemma}
where \( D \) is the dimension of the covariance matrix. We define matrix entropy on the token matrix \(\mathbf{X}\) and provide the proof in Appendix~\ref{Proofs}. 

To provide an intuition of its role in the token matrix, we introduce effective rank~\citep{roy2007effective}, which links matrix entropy to dimensionality. For the specific definition, please refer to Appendix~\ref{Proofs}. Recent works~\citep{zhang2023matrix, zhuo2023towards} investigate the relationship between matrix entropy and effective rank. Inspired by these studies, we adopt effective rank to quantify the effective information across heads and layers to explore the model's information compression patterns. Based on these patterns, we set different compression ratios for different heads and layers. We define the compression ratio \(\rho\) for each \(\mathbf{X}\) as the ratio between the compressed length \(\hat{N}\) and the original sequence length \(N\), i.e.,
 \begin{equation}
\small
\label{compression rate}
\rho = \frac{\hat{N}}{N}.
\end{equation}%  

\subsection{Truncated Matrix Entropy}
\begin{figure}[ht]
% \vspace{-3mm}
\centering
\includegraphics[width=\columnwidth]{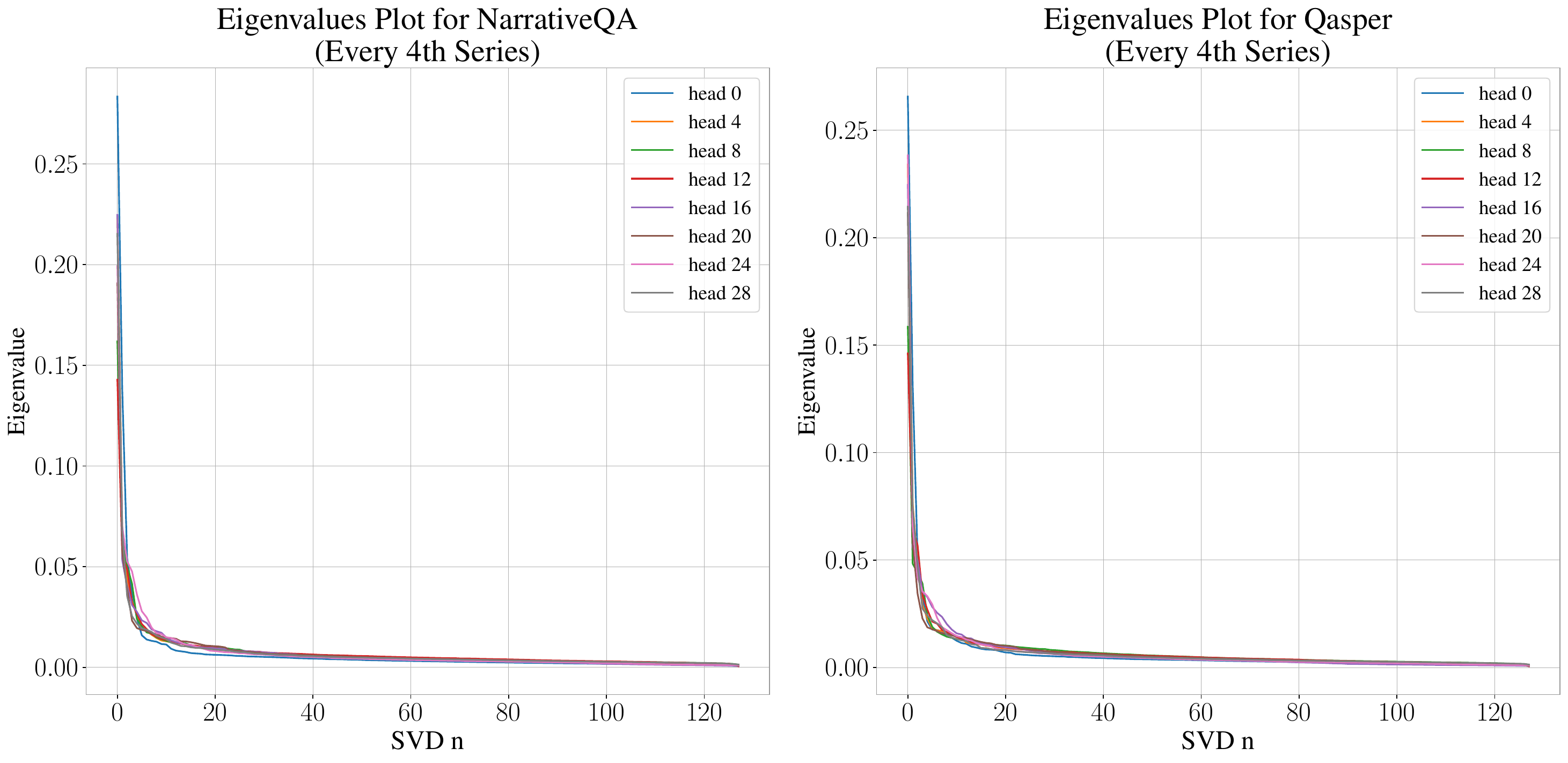}
\caption{The spectrum of \(\mathbf{\Sigma}_{{Q_m}}\) across two datasets in LongBench and various heads.}
% \vspace{-3mm}
\label{fig:eigenvalue_distribution}
\end{figure}
\begin{figure}[ht]
\centering
\includegraphics[width=\columnwidth]{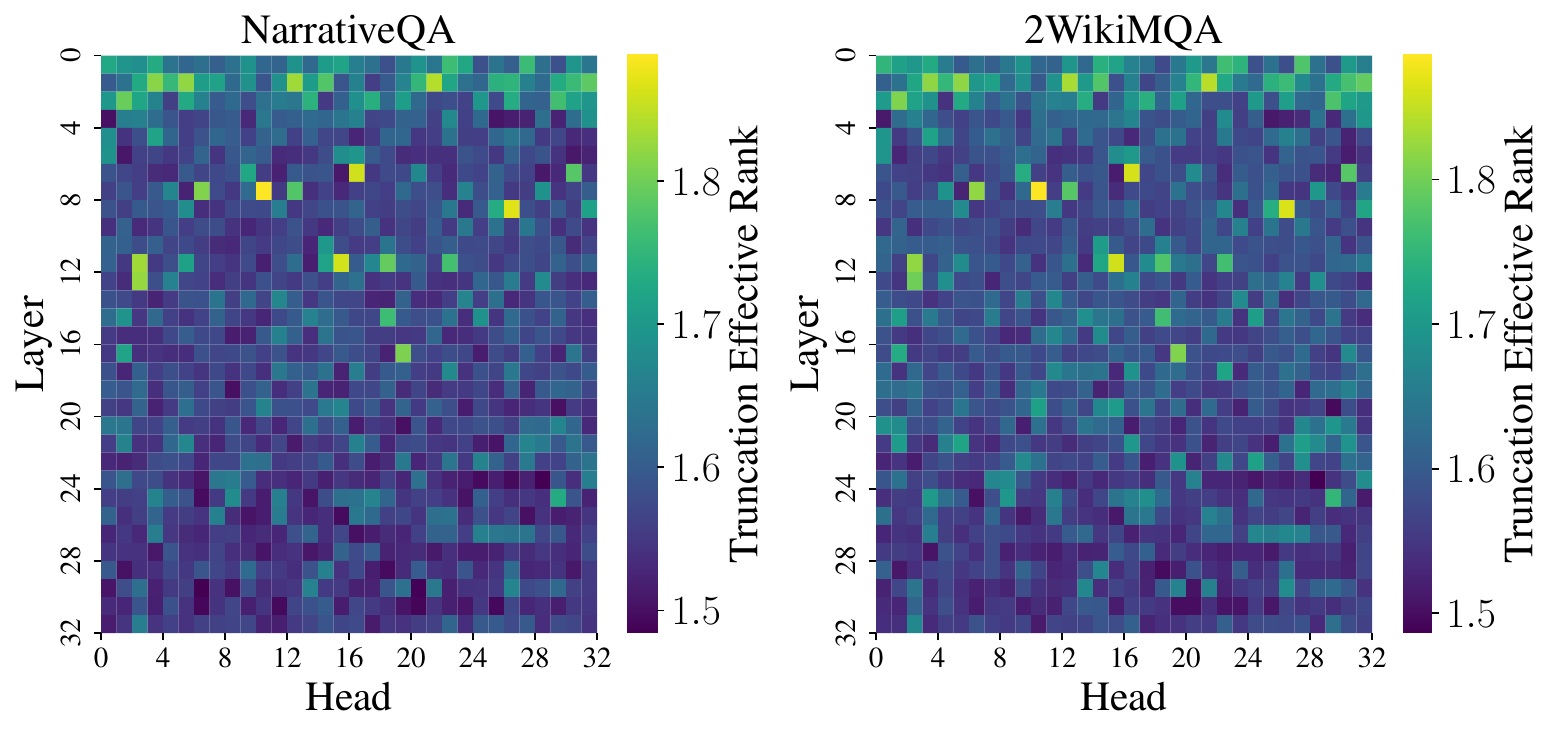}
% \vspace{-8mm}
\caption{The heatmap of \( \text{erank}_{k}(\mathbf{\Sigma}_{Q_m}) \) across different layers and heads.}
% \vspace{-2mm}
\label{fig:eigenvalue_heatmap_distribution1}
\end{figure}

To uncover \textit{information compression patterns} in the token matrix, how do we determine which matrix—key (\( K_m \)), query (\( Q_m \)), value (\( V_m \)), or hidden state (\( H_m \))—best captures the compression patterns in the token matrices? To answer this question, we propose \textit{truncated matrix entropy}.

% \vspace{-2mm}
\paragraph{Observation} To gain insight, we first focus on the \textit{spectrum} of \(\mathbf{\Sigma}_{{Q_m}}\)~\citep{tang2024quest} across different heads in the model's final layer. Figure~\ref{fig:eigenvalue_distribution} reveals: \textit{i)} The initial part of the eigenvalue distribution varies significantly, suggesting that only a small portion of the feature components are active. \textit{ii}) Eigenvalue distributions across heads differ significantly in the leading part, prompting us to use this part to calculate the model's effective rank and distinguish head types. 

A key observation from Figure~\ref{fig:eigenvalue_heatmap_distribution1} is that, in different layers, there are heads with abnormally high entropy, with most of them found in the middle layers (9-15). Later empirical experiments show that these high-entropy heads and their corresponding layers are associated with special sparsity patterns that have long-range dependency (retrieval layers). 

% \vspace{-2mm}
\paragraph{Discussion} \textit{By Lemma~\ref{eq:Lemma 3.2.}, the covariance matrix used for computing the matrix entropy must be positive definite.} Therefore, we calculate the effective rank of \(\mathbf{\Sigma}_{Q_m}\) using a positive definite submatrix. Furthermore, as shown in Figure~\ref{fig:eigenvalue_distribution}, we observe a distinct elbow point~\citep{thorndike1953belongs}, where the eigencomponents preceding the elbow point represent the principal components of the matrix.

% \vspace{-1mm}
 Based on the above observations, we select the top-$k$ eigenvalues before the elbow point~\citep{kaiser1960application} and compute the effective rank. This leads to the definition of \textit{truncated effective rank}, which quantify uncertainty through the effective rank of a submatrix:
 % \vspace{-2mm}
\begin{equation}
\small
\label{eq:eigenvalues_entropy}
H_{k}(\mathbf{\Sigma}_{\mathbf{X}}) = - \sum_{i=1}^{k} \sigma_i \log \sigma_i,
\end{equation}
%\vspace{-5mm}
\begin{equation}
\small
\text{erank}_{k}(\mathbf{\Sigma}_{\mathbf{X}}) = \exp \left( H_{{k}}\left(\mathbf{\Sigma}_{\mathbf{X}}\right) \right).
\end{equation}
With \( \text{erank}_{k}(\mathbf{\Sigma}_{\mathbf{X}}) \) defined, we begin classifying the compression patterns and exploring the estimation of the compression ratio of \( H_m \) across layers and \( K_m \), \( V_m \) across heads.

% \textcolor{blue}{how can we leverage \( \text{erank}_{k}(\mathbf{\Sigma}_{Q_m}) \) to estimate rate for $H_m$, $K_m$, and $V_m$? what's their connection}

\begin{figure}[t]
\centering
% \vspace{-1mm}
\includegraphics[width=\columnwidth]
{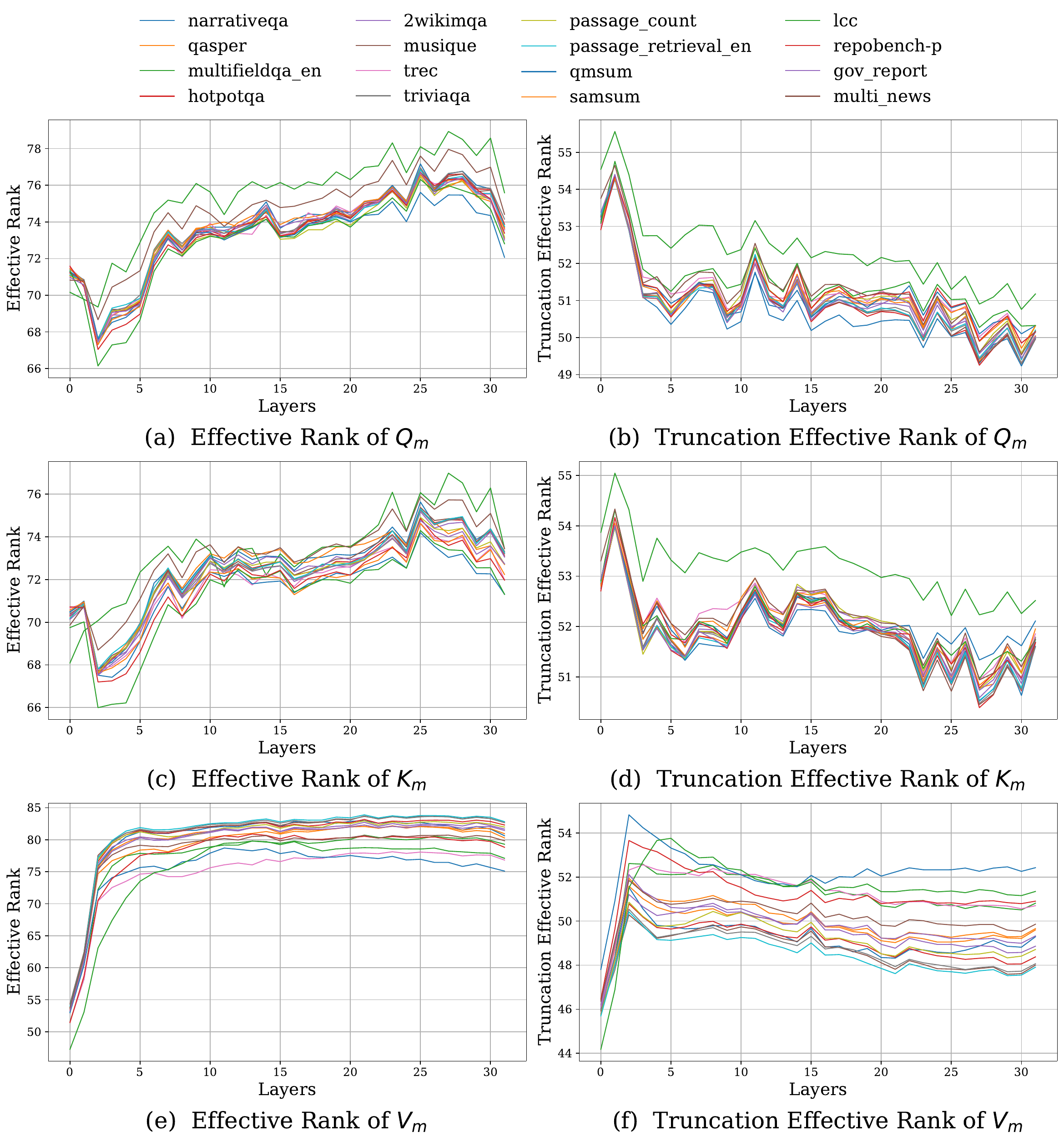}
% \vspace{-8mm}
\caption{Effective rank and truncated effective rank for the \( Q_m \), \( K_m \), and \( V_m \) across different layers.}
\label{fig:matrix_entropy_query_key_value}
% \vspace{-5mm}
\end{figure}

% \vspace{-2mm}
% \paragraph{Observation} We plot the entropy variation trends for \( Q_m \), \( K_m \), and \( V_m \) across different layers and datasets in Figure~\ref{fig:matrix_entropy_query_key_value}. Figure~\ref{fig:matrix_entropy_query_key_value} reveals: \textit{i}) \( Q_m \) and \( K_m \) exhibit a more pronounced entropy increase or decrease trend than \( V_m \) (\( V_m \) and \( H_m \) show similar compression patterns, see Appendix~\ref{Implementation details}), indicating that \( Q_m \) and \( K_m \) serve as stronger indicators of information compression patterns compared to \( V_m \) and \( H_m \). \textit{ii}) Truncated effective rank and effective rank exhibit distinctly different variation trends, especially in \( Q_m \) and \( K_m \), \textit{revealing the fundamentally different information compression behaviors between the two}. \textit{iii}) As the model depth increases, inter-layer \(\text{erank}_{k}(\mathbf{\Sigma}_{Q_m})\) and \(\text{erank}_{k}(\mathbf{\Sigma}_{K_m})\) show entropy reduction, indicating sparser patterns~\citep{wang2023label}. \textit{iv}) The truncated matrix entropy of \( Q_m \) shows a greater reduction in entropy compared to \( K_m \) from the initial layer to the final layer, with a lower average entropy, suggesting that it is an effective indicator. 
% \vspace{-2mm}
% \paragraph{Discussion} Based on the above observations, to bridge the gap between information compression patterns and sparsity patterns, we choose \( Q_m \) to estimate the sparsity patterns of \( K_m \), \( V_m \), and \( H_m \).

\paragraph{Observation} 
We visualize the entropy variation trends of \( Q_m \), \( K_m \), and \( V_m \) across different layers and datasets in Figure~\ref{fig:matrix_entropy_query_key_value}. The figure reveals the following: \textit{i}) \( Q_m \) and \( K_m \) exhibit more pronounced trends of entropy increase or decrease compared to \( V_m \) (notably, \( V_m \) and \( H_m \) demonstrate similar compression patterns; see Appendix~\ref{Implementation details}), suggesting that \( Q_m \) and \( K_m \) serve as stronger indicators of information compression than \( V_m \) and \( H_m \). \textit{ii}) \textit{Truncated effective rank} and full effective rank display markedly different variation patterns, particularly in \( Q_m \) and \( K_m \), \textit{highlighting fundamentally different behaviors in information compression}. \textit{iii}) As model depth increases, the inter-layer \(\text{erank}_{k}(\mathbf{\Sigma}_{Q_m})\) and \(\text{erank}_{k}(\mathbf{\Sigma}_{K_m})\) show a decreasing trend, indicating increasingly sparse structures~\citep{wang2023label}. \textit{iv}) From the initial to the final layer, the \textit{truncated matrix entropy} of \( Q_m \) decreases more significantly than that of \( K_m \), and its average entropy is also lower, reinforcing its role as an effective indicator.

% \vspace{-2mm}

\paragraph{Discussion} 
Based on the above observations, to better connect information compression patterns with sparsity patterns, we select \( Q_m \) as a proxy to estimate the sparsity characteristics of \( K_m \), \( V_m \), and \( H_m \) and design the two-stage compression strategy presented in the next section.

% \vspace{-2mm}
\subsection{Uncertainty-Aware Compression Strategy}
\begin{figure*}[htbp]
    \centering
    \includegraphics[width=\textwidth]{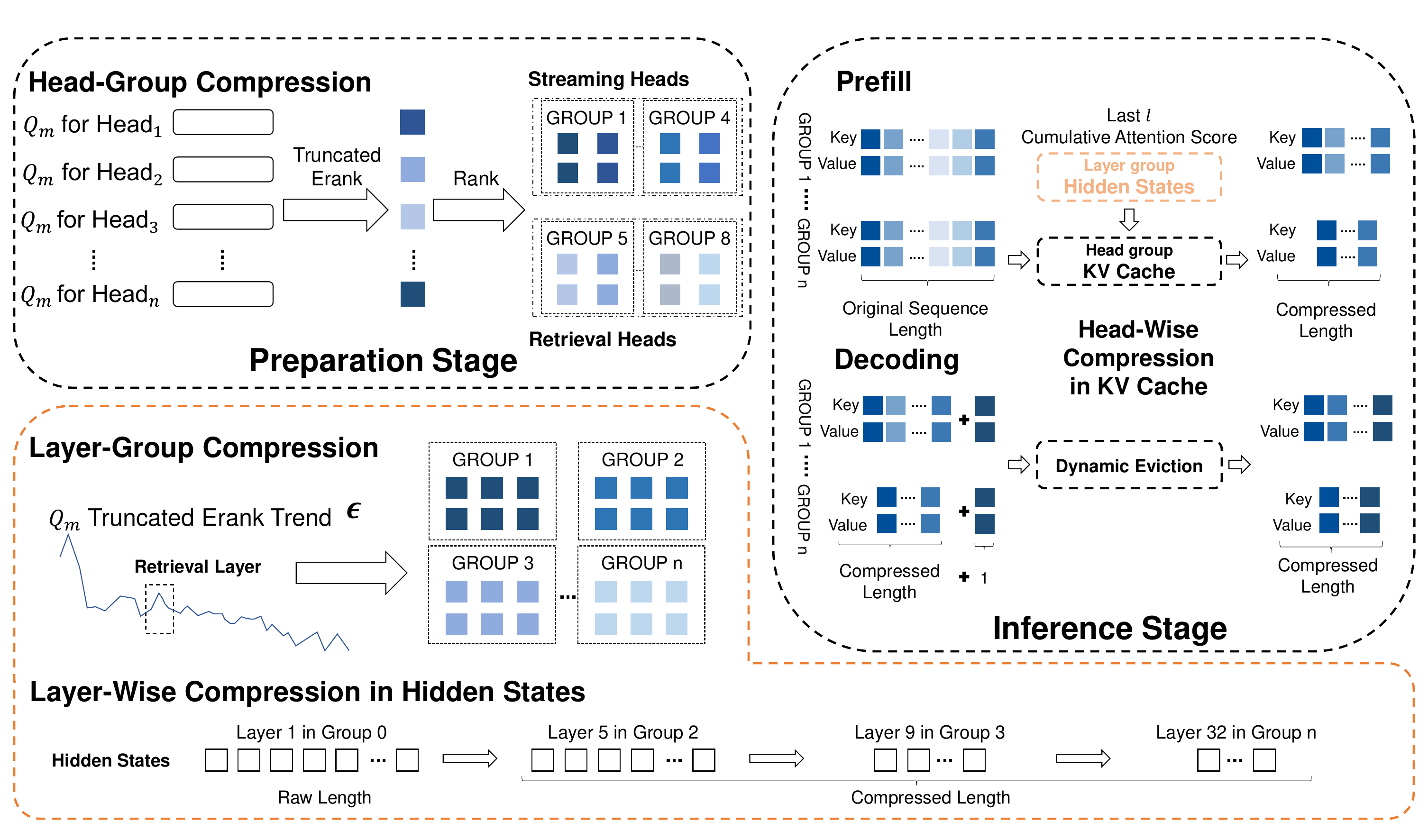} 
    % 图片路径
    % \includegraphics[width=\textwidth]
    % {figure/follow_new.pdf}
    \caption{Overview of \textsc{UNComp}. Darker colors indicate more retained hidden states and a weaker compression (i.e., a higher $\rho$) corresponding to higher attention scores.}
    \label{fig:model}
    % \vspace{-6mm}
\end{figure*}
In this section, we provide a detailed explanation of how \textit{truncated matrix entropy} is used to guide compression. We first introduce the preparation stage, followed by a two-stage compression strategy, along with the mapping between information compression and sparsity patterns. The workflow of our method is illustrated in Figure~\ref{fig:model}.
%\vspace{-2mm}
% \vspace{-1mm}
\subsubsection{Preparation Stage}
\paragraph{Observation} We first observe in Figure~\ref{fig:eigenvalue_heatmap_distribution1} that attention head information compression patterns remain consistent across datasets, suggesting that the sparsity pattern of heads is not data-dependent.

% \vspace{-3mm}
\paragraph{Design} Since the head information compression pattern is not data-dependent, we first sample 500 data points from Wikitext2~\citep{merity2016wikitext} before inference to pre-group the heads. These samples are used to group the model heads and set compression ratios for each group, which helps identify the retrieval heads in the KV cache. Specifically, after inputting \(d = 500\) data points into the model, we save the KV cache of all data and calculate the \(\text{erank}_{k}(\mathbf{\Sigma}_{Q_m}^{(i, h)})\) for each head of each data point. Then, we average it to obtain \(\hat{\text{erank}}_{k}(\mathbf{\Sigma}_{Q_m}^{h})\) as the grouping metric for the inference stage:
\begin{equation}
\small
\label{eq:Preparation Stage}
\hat{\text{erank}}_{k}(\mathbf{\Sigma}_{Q_m}^{h}) = \frac{1}{d} \sum_{i=1}^{d} \text{erank}_{k}(\mathbf{\Sigma}_{Q_m}^{(i,h)}),
\end{equation}
where \(i\) is the \(i\)-th data point, and \(h\) is the \(h\)-th head. 
 
%\vspace{-2mm}

\subsubsection{Layer-Group Compression}
For the first-stage compression, we focus on compressing the hidden states \(H_m\) and attempt to identify retrieval-layer. 
% \vspace{-1mm}
\paragraph{Observation} Previous KV cache compression methods~\citep{zhang2024pyramidKV, zhang2024h2o} do not reduce computation, as they evict the KV cache only after prefill. In contrast, we compress hidden states before generating the KV cache, saving both computation and memory. From Figure~\ref{fig:matrix_entropy_query_key_value}(b), we observe that some layers show an increase in entropy, indicating that the token matrix is unsuitable for compression as it aggregates information.

 % \vspace{-2mm}
\paragraph{Design} Specifically, we divide the \(L\) layers into \(C\) groups, ensuring that the token length within each group remains consistent across layers. By applying the compression patterns in Figure~\ref{fig:matrix_entropy_query_key_value}(a), we compress layers where inter-layer entropy reduction \(\Delta\) \(c_i\) falls below a threshold \( \epsilon \), determining the total compression stages \( C \):
\begin{equation}
\small
\label{eq:stage1}
\Delta c_i =   \text{erank}_{k}(\mathbf{\Sigma}_{Q_m}^{i}) - \text{erank}_{k}(\mathbf{\Sigma}_{Q_m}^{i+1}) ,
\end{equation}
\begin{equation}
\small
\label{eq:stage2}
C = \sum_{i=1}^{L-1} \mathbf{1} ( \Delta c_i > \epsilon > 0 ),
\end{equation}
where \( \mathbf{1} \) is the indicator function that evaluates to 1 if the entropy reduction between layer \( i \) and \( i+1 \) exceeds the threshold \( \epsilon \), and 0 otherwise. Eq.~\eqref{eq:stage1} is a partition function determining the division of the model's layers into \(C\) groups. The context size at each subsequent group is calculated by%
% \vspace{-1mm}
\begin{equation}
\small
\label{eq:token_number}
N_{i+1} = N_i + \Delta n, \quad i = 1, 2, \dots, C-1,
\end{equation}
where \( \Delta n \) (hyperparameter) represents the increment between consecutive groups. With the token budget \(N_i\) for each group \(i\) determined, hidden state eviction starts from the second group, while the first group's hidden states remain full-size to preserve information in early layers.
% \vspace{-2mm}

\begin{table*}[t]
\vspace{-1mm}
\centering
\adjustbox{max width=\textwidth}{
\scriptsize
\begin{tabular}{c@{}c@{}c@{}c@{} c@{}c@{}c@{} c@{}c@{}c@{} c@{}c@{}c@{} c@{}c@{} c@{}c@{} c@{} c}
\toprule
\multirow{2}{*}{\raisebox{-4ex}{\textbf{Methods}}}  % Moves the text down
& \multicolumn{3}{c}{\textbf{Single-Document QA}} 
& \multicolumn{3}{c}{\textbf{Multi-Document QA}} 
& \multicolumn{3}{c}{\textbf{Summarization}} 
& \multicolumn{3}{c}{\textbf{Few-shot Learning}} 
& \multicolumn{2}{c}{\textbf{Synthetic}} 
& \multicolumn{2}{c}{\textbf{Code}} 
& \multirow{2}{*}{\raisebox{-4ex}{\textbf{Avg.}}} 
% & \multirow{2}{*}{\raisebox{-4ex}{\textbf{ Time}}} 
& \multirow{2}{*}{\raisebox{-6ex}{\textbf{\shortstack{Time \\ (s / sample)}}}}
\\  % Moves the text down

\cmidrule(lr){2-4} \cmidrule(lr){5-7} \cmidrule(lr){8-10} \cmidrule(lr){11-13} \cmidrule(lr){14-15} \cmidrule(lr){16-17}
\setlength{\tabcolsep}{1pt} % 默认值是6pt，减小这个值来减少列间距
& \makebox[1cm]{\raisebox{0.5ex}{\rotatebox{30}{\textbf{NtrvQA}}}} 
& \makebox[1cm]{\raisebox{0.7ex}{\rotatebox{30}{\textbf{Qasper}}}} 
& \makebox[1cm]{\raisebox{0.8ex}{\rotatebox{30}{\textbf{MF-en}}}} 
& \makebox[1cm]{\raisebox{0.4ex}{\rotatebox{30}{\textbf{HotpotQA}}}} 
& \makebox[1cm]{\raisebox{0.3ex}{\rotatebox{30}{\textbf{2WikiMQA}}}} 
& \makebox[1cm]{\raisebox{0.7ex}{\rotatebox{30}{\textbf{Musique}}}} 
& \makebox[1cm]{\raisebox{0.5ex}{\rotatebox{30}{\textbf{GovReport}}}} 
& \makebox[1cm]{\raisebox{0.8ex}{\rotatebox{30}{\textbf{QMSum}}}} 
& \makebox[1cm]{\raisebox{0.6ex}{\rotatebox{30}{\textbf{MultiNews}}}} 
& \makebox[1cm]{\raisebox{0.8ex}{\rotatebox{30}{\textbf{TREC}}}} 
& \makebox[1cm]{\raisebox{0.6ex}{\rotatebox{30}{\textbf{TriviaQA}}}} 
& \makebox[1cm]{\raisebox{0.6ex}{\rotatebox{30}{\textbf{SAMSum}}}} 
& \makebox[1cm]{\raisebox{0.6ex}{\rotatebox{30}{\textbf{PCount}}}}  
& \makebox[1cm]{\raisebox{1.4ex}{\rotatebox{30}{\textbf{PRe}}}}    
& \makebox[1cm]{\raisebox{1.6ex}{\rotatebox{30}{\textbf{Lcc}}}} 
& \makebox[1cm]{\raisebox{1.4ex}{\rotatebox{30}{\textbf{RB-P}}}} \\

% Llama-2-7B-chat-hf
\midrule
\multicolumn{19}{c}{\textbf{Llama2-7B-chat-hf, KV Size = FULL}} \\
\arrayrulecolor[gray]{0.8}
\midrule
\arrayrulecolor{black}
% \rowcolor{lightgray} 
FullKV & 19.34 & 18.61 & 35.19 & 30.66 & 28.42 & 10.05 & 25.19 & 20.18 & 25.73 & 63.00 & 83.62 & 41.60 & 5.00 & 10.00 & 61.40 & 55.45 & 33.34 & 0.96 \\
\midrule
\multicolumn{18}{c}{\textbf{Llama-2-7B-chat-hf, KV Size = 384 , Compressibility is 9.38\%  (Except CHAI method)}} \\
\arrayrulecolor[gray]{0.8}
\midrule
\arrayrulecolor{black}
H2O & 14.96 & 14.60 & 17.40 & 26.72 & 27.97 & 6.11 & 17.83 & 18.76 & 20.17 & 47.00 & 77.56 & 39.39 & 4.50 & 5.00 & 57.08 & 50.31 & 27.84 & 0.94  \\
StreamingLLM & 13.71 & 13.68 & 19.40 & 26.97 & 28.03 & 6.78 & 15.13 & 18.87 & 18.27 & 46.50 & 80.02 & \textbf{40.85} & 4.50 & 5.00 & 56.84 & 51.56 & 27.88 & 0.88 \\
SnapKV & 16.27 & 17.34 & 30.37 & \textbf{33.04} & 27.82 & 9.92 & 19.34 & 20.33 & 22.63 & 59.50 & 83.50 & 38.45 & \textbf{5.50} & \textbf{12.50} & 59.18 & \textbf{55.28} & 31.94 & 0.82\\
PyramidKV & 16.86 & 18.26 & 31.01 & 31.59 & 27.93 & 8.69 & 19.88 & 20.15 & 22.43 & 62.00 & 83.86 & 38.98 & \textbf{5.50} & 10.00 & 58.94 & 52.80 & 31.81 & 0.84 \\
CHAI & 16.75 & 16.91 & \textbf{34.69} & 26.09 & 20.80 & 9.20 & 20.79 & 20.23 & 23.33 & 57.00 & 75.52 & 35.67 & 4.00 & 6.33 & 50.10 & 46.55 & 29.00 & 1.51 \\
Quest & 17.31 & 19.55 & 32.18 & 30.25 & 27.20 & 9.48  & 22.82 & 19.25 & \textbf{25.99} & 62.50 & 83.26 & 40.37 & 5.00 & 5.25  & 58.81 & 53.24 & 32.03 & 0.96 \\
Double-Sparse & 17.27 & 19.85 & 32.30 & 29.45 & \textbf{28.54} & \textbf{9.90}  & 20.88 & 19.84 & 25.38 & 61.50 & 83.48 & 40.56 & 5.25 & 8.00  & 52.27 & 51.97 & 31.65 & 1.02 \\ 
RazorAttention & 16.88 & 19.33 & 32.44 & 31.28 & 27.81 & 9.46 & \textbf{24.29} & 20.02 & 23.19 & 61.00 & 83.85 & 40.28 & 5.00 & 12.00 & 54.71 & 52.29 & 32.11 & 1.21 \\
Ours-group-stage & \textbf{17.70} & \textbf{20.28} & 30.98 & 30.35 & 27.37 & 9.29 & 22.25 & \textbf{20.60} & 24.04 & \textbf{63.00} & 83.68 & 38.27 & 4.50 & 6.50 & 58.02 & 53.79 & 31.91 & \textbf{0.57} \\
Ours-group & 17.12 & 19.20 & 32.09 & 30.99 & 27.88 & 9.27 & 23.20 & 20.10 & 24.72 & 62.50 & \textbf{84.72} & 39.33 & \textbf{5.50} & 11.10 & \textbf{59.71} & 54.15 & \textbf{32.60} & 0.83 \\

% Llama-2-13B-chat-hf
\midrule
\multicolumn{18}{c}{\textbf{Llama2-13B-chat-hf, KV Size = FULL}} \\
\arrayrulecolor[gray]{0.8}
\midrule
\arrayrulecolor{black}
% \rowcolor{lightgray} 
FullKV & 18.20 & 26.07 & 37.06 & 36.20 & 32.44 & 14.19 & 25.82 & 20.20 & 26.00 & 66.50 & 87.49 & 35.93 & 3.12 & 11.50 & 53.29 & 52.73 & 34.17 & 2.21 \\
\midrule
\multicolumn{18}{c}{\textbf{Llama-2-13B-chat-hf, KV Size = 384 , Compressibility is 9.38\%  (Except CHAI method)}} \\
\arrayrulecolor[gray]{0.8}
\midrule
\arrayrulecolor{black}
H2O & 14.11 & 18.36 & 22.78 & 33.03 & 27.58 & 12.94 & 18.97 & 18.69 & 20.37 & 53.50 & 85.75 & 34.15 & 3.55 & 6.00 & 50.97 & 47.56 & 29.27 & 2.57 \\
StreamingLLM & 13.23 & 18.47 & 23.76 & 34.50 & 29.62 & 11.09 & 18.67 & 18.47 & 17.89 & 52.50 & 84.93 & 32.54 & 3.55 & 7.00 & 40.60 & 42.86 & 28.11 & 1.96\\
SnapKV & 17.09 & 22.77 & 34.37 & 36.73 & 31.04 & 13.02 & 19.70 & 20.00 & 22.91 & 62.00 & 87.48 & \textbf{37.44} & \textbf{4.05} & 11.50 & 51.76 & \textbf{51.27} & 32.70 & 1.93 \\
PyramidKV & 16.33 & 22.81 & 34.19 & \textbf{37.54} & 30.25 & 13.82 & 19.79 & 20.11 & 23.14 & 64.50 & 86.45 & 36.62 & \textbf{4.05} & \textbf{12.00} & 52.06 & 50.58 & 32.77 & 3.50 \\
CHAI & 17.06 & 23.51 & 31.01 & 33.70 & 27.78 & 11.73 & 23.03 & 19.59 & 24.66 & 65.00 & 86.18 & 15.93 & 4.00 & 8.50 & 45.57 & 48.74 & 30.37 & 2.50 \\
Quest & 17.07 & \textbf{26.36} & 34.56 & 34.50 & 29.62 & 13.19 & 23.79 & 19.71 & 25.32 & 64.00 & 84.93 & 36.46 & 2.62 & 9.50 & \textbf{52.12} & 49.85 & 32.73 & 1.86 \\
Double-Sparse & 17.42 & 25.10 & 31.85 & 33.27 & 29.89 & 12.61 & 24.00 & 19.88 & \textbf{25.68} & 67.00 & 86.48 & 35.97 & 2.85 & 14.31 & 51.95 & \textbf{51.27} & 33.10 & 1.98 \\
RazorAttention & 17.28 & 25.43 & 36.88 & 35.27 & \textbf{30.11} & 12.89 & \textbf{25.02} & 20.17 & 24.98 & 65.00 & 84.29 & 35.28 & 3.81 & 11.00 & 51.89 & 49.29 & 33.04 & 2.44 \\
Ours-group-stage & 16.02 & 23.90 & 35.19 & 36.66 & 30.50 & \textbf{14.19} & 19.71 & 19.69 & 23.78 & 62.00 & 86.04 & 35.91 & 3.55 & 8.50 & 49.37 & 46.73 & 31.98  & \textbf{1.32} \\
Ours-group & \textbf{18.27} & 24.33 & \textbf{37.12} & 36.46 & 29.83 & 13.57 & 21.15 & \textbf{20.84} & 24.07 & \textbf{67.50} & \textbf{87.96} & 36.42 & 3.55 & \textbf{12.00} & 51.50 & 50.72 & \textbf{33.46} & 1.95 \\

%Llama-3-8B-Instruct
\midrule
\multicolumn{18}{c}{\textbf{Llama3-8B-Instruct, KV Size = FULL}} \\
\arrayrulecolor[gray]{0.8}
\midrule
\arrayrulecolor{black}
FullKV & 23.31 & 31.18 & 38.09 & 43.67 & 35.26 & 21.43 & 28.42 & 22.90 & 26.64 & 73.50 & 89.76 & 42.20 & 4.78 & 67.88 & 60.12 & 56.76 & 41.62 & 2.98\\
\midrule
\multicolumn{18}{c}{\textbf{Llama-3-8B-Instruct, KV Size = 384 , Compressibility is 4.74\%  (Except CHAI method)}} \\
\arrayrulecolor[gray]{0.8}
\midrule
\arrayrulecolor{black}
H2O & 18.80 & 13.76 & 21.20 & 38.90 & 31.38 & 14.81 & 20.38 & 20.70 & 22.03 & 61.00 & 82.07 & 39.49 & 5.12 & 66.92 & 58.59 & 54.98 & 35.63 & 3.98 \\
StreamingLLM & 19.39 & 10.44 & 21.98 & 39.39 & 30.03 & 14.29 & 20.37 & 20.82 & 22.62 & 57.50 & 82.89 & 39.73 & 5.25  & 68.00 & 58.68 & 55.67 & 35.44 & 2.78\\
SnapKV & 21.47 & 19.77 & 33.97 & 43.10 & 32.79 & \textbf{21.48} & 21.69 & 22.01 & 22.92 & 63.00 & 89.69 & 39.78 & 5.06 & 67.83 & 60.19 & 56.82 & 38.85 & 2.68 \\
PyramidKV & 22.08 & 19.43 & 32.99 & 42.51 & 32.01 & 19.62 & 21.73 & 22.24 & 22.74 & 71.00 & 89.59 & 40.51 & 4.23 & \textbf{68.50} & 58.92 & 53.92 & 38.88 & 2.78 \\
CHAI & 18.99 & 23.44 & 31.82 & 33.37 & 22.63 & 19.07 & 24.46 & 21.74 & 23.78 & 69.00 & 89.28 & 37.15 & 4.92 & 67.75 & 44.44 & 36.12 & 35.50 & 4.20 \\
Quest & 21.35 & 23.40 & 34.56 & 32.94 & 31.34 & 14.93 & 19.68 & 22.18 & 26.07 & 71.00 & 88.30 & \textbf{41.32} & 5.37 & 67.06 & 54.26 & 46.25 & 37.50 & 2.84\\
Double-Sparse & 21.92 & \textbf{29.86} & 33.09 & 27.85 & 29.49 & 11.11 & \textbf{27.89} & 21.41 & \textbf{26.66} & 71.00 & 88.52 & 40.80 & \textbf{5.68} & 65.55 & 57.18 & 54.77 & 38.30 & 2.98 \\
RazorAttention & \textbf{22.36} & 26.85 & 34.28 & 42.87 & 34.28 & 20.85 & 24.28 & 21.28 & 24.98 & 69.00 & 89.28 & 40.28 & 4.28 & 67.00 & 58.28 & 52.87 & 39.56 & 3.02 \\
Ours-group-stage  & 21.32 & 26.42 & 33.98 & \textbf{43.18} & \textbf{35.38} & 19.76 & 22.47 & 22.13 & 22.89 & \textbf{72.00} & 90.26 & 39.29 & 4.81 & 60.50 & \textbf{61.97} & 57.61 & 39.62 & \textbf{1.79}\\
Ours-group & 22.27 & 26.57 & \textbf{36.04} & \textbf{43.18} & 35.25 & 20.57 & 22.71 & \textbf{22.31} & 22.80 & \textbf{72.00} & \textbf{90.71} & 40.59 & 5.21 & 68.00 & 61.85 & \textbf{58.31} & \textbf{40.52} & 2.70  \\

\midrule
\multicolumn{18}{c}{\textbf{Qwen2.5-7B-Instruct, KV Size = 8k}} \\
\arrayrulecolor[gray]{0.8}
\midrule
\arrayrulecolor{black}
% \rowcolor{lightgray} 
FullKV & 23.64 & 40.50 & 50.03 & 43.44 & 46.28 & 26.67 & 32.94 & 22.50 & 25.32 & 69.00 & 88.22 & 39.87 & 3.72 & 63.00 & 6.70 & 3.93 & 36.61 & 2.45 \\
\midrule
\multicolumn{18}{c}{\textbf{Qwen2.5-7B-Instruct, KV Size = 384 , Compressibility is 9.38\%  (Except CHAI method)}} \\
\arrayrulecolor[gray]{0.8}
\midrule
\arrayrulecolor{black}
H2O & 20.62 & 23.90 & 27.84 & 36.66 & 34.71 & 19.84 & 25.49 & 19.64 & 21.67 & 39.50 & 82.55 & \textbf{37.89} & 5.25 & 16.50 & 8.33 & \textbf{10.96} & 26.96 & 3.14 \\
StreamingLLM & 17.20 & 21.26 & 32.74 & 37.54 & 36.97 & 18.64 & 25.14 & 18.81 & 23.54 & 45.00 & 73.39 & 32.10 & \textbf{6.58} & 16.50 & \textbf{9.10} & 7.29 & 26.36 & 2.64 \\
SnapKV  & 24.16 & 36.32 & 47.64 & 42.64 & 42.60 & 23.42 & 25.64 & 20.87 & 21.64 & 65.00 & 87.45 & 37.65 & 5.37 & 69.50 & 5.78 & 6.13 & 35.11 & 2.63 \\
PyramidKV  & 21.72 & 34.32 & 45.25 & \textbf{44.43} & 41.43 & \textbf{24.93} & 23.72 & 20.51 & 20.18 & 59.50 & 87.09 & 36.49 & 4.68 & 69.00 & 5.68 & 5.84 & 34.05 & 2.61 \\
CHAI & 22.98 & 38.42 & 42.68 & 43.69 & 39.65 & 22.58 & 24.58 & 20.64 & 21.87 & \textbf{66.00} & 86.98 & 36.97 & 4.89 & 69.00 & 3.64 & 5.64 & 34.39 & 3.48 \\
Quest & 24.36 & 36.84 & 48.25 & 42.65 & 40.80 & 24.68 & 23.54 & 20.23 & 22.58 & 65.00 & 86.54 & 37.61 & 4.89 & 67.00 & 8.12 & 8.26 & 35.08 & 2.89 \\
Double-Sparse & 23.56 & 37.24 & \textbf{48.64} & 43.54 & 41.26 & 24.54 & \textbf{30.29} & 20.41 & \textbf{23.43} & \textbf{66.00} & 87.65 & 36.54 & 4.36 & 68.00 & 6.34 & 5.64 & 35.47 & 2.76 \\
RazorAttention & 21.45 & 35.68 & 47.29 & 42.76 & 39.86 & 24.86 & 24.98 & 20.18 & 21.28 & {66.00} & 88.24 & 37.51 & 5.89 & 70.00 & 6.12 & 8.24 & 35.02 & 2.88 \\
Ours-group-stage  & \textbf{25.13} & 40.22 & 48.49 & 44.21 & 40.67 & 24.75 & 26.05 & \textbf{21.37} & 22.04 & 65.00 & 88.47 & 37.76 & 5.05 & \textbf{70.50} & 8.44 & 6.37 & \textbf{35.91} & \textbf{1.64} \\
Ours-group & 23.91 & \textbf{40.31} & 48.41 & 43.97 & \textbf{41.89} & 23.81 & 25.89 & 21.10 & 21.91 & \textbf{66.00} & \textbf{89.26} & 37.59 & 4.69 & 70.00 & 8.38 & 6.20 & 35.83 &2.71 \\

% & &  & \textbf{48.77} & 36.80 & 24.83 & 15.22 & 24.87 & 21.65 & 23.78 & 69.50 & 85.93 & 43.17 & 0.94 & 27.38 & 55.28 & 50.91 & 35.71 \\

\bottomrule
\end{tabular}
}
\vspace{-2mm}
\caption{Performance Comparison across Different Tasks: Ours-group-stage compresses both hidden states and KV cache, while Ours-group compresses only the KV cache. Ours-group-stage is 68.42\% faster than FullKV, with only a 1.43\% performance loss. All methods compress key and value caches at the same compression ratio.}
\vspace{-4mm}
\label{main-table}

\end{table*}

\paragraph{Hidden State Eviction}  We observe that \( H^{(i)}_m \) and \( V^{(i)}_m \) share the same information compression pattern. Based on this observation, we decide to use the attention scores to determine the eviction strategy for \( H^{(i)}_m \). From group 2 to group \(C\), we predict token eviction in \( H^{(i+1)}_m \) using the attention scores of the last token in the \(i\)th layer, retaining the tokens with the highest \(N_{i+1}\) attention scores. After generating \( H^{(i+1)}_m \), we map \( H^{(i+1)}_m \) to the three matrices \( Q^{(i+1)}_m \), \( K^{(i+1)}_m \), and \( V^{(i+1)}_m \). Please refer to Appendix~\ref{algo} for more details. 

% \vspace{-2mm}
\paragraph{Retrieval Layer}
We select our retrieval layers based on the maximum entropy increase layer and use average interpolation to merge it with the parameters of the final layer to improve performance.
% \vspace{-5mm}
 \begin{table}[ht]
    \centering
    \adjustbox{max width=\textwidth}{%
    \scriptsize
    \begin{tabular}{c c c c}
    \toprule
    \multirow{1}{*}{{\textbf{Dataset}}}  
    & \multirow{1}{*}{{\textbf{High Entropy}}} 
    & \multirow{1}{*}{{\textbf{Random Group}}} 
    & \multirow{1}{*}{{\textbf{Low Entropy}}} \\
    \midrule
    Qasper & 15.40 & 11.70 & 10.60   \\
    Musique  & 6.96 & 4.47 & 4.50 \\ 
    GovReport & 14.88 & 7.31 & 3.20   \\ 
    TREC & 55.50 & 32.50 & 30.45  \\ 
    PCount & 5.00 & 3.97 & 3.90  \\
    Lcc  & 50.00 & 32.50 & 30.55  \\
    Average & \textbf{24.62} & 15.49 & 13.86 \\
    \bottomrule
    \end{tabular}
    }
    \vspace{-2mm}
    \caption{Head Sparsity Patterns.}
    \label{random_experiment}
    \vspace{-7mm}
\end{table}

% \vspace{-1mm}
\subsubsection{Head-Group Compression}
After the prefill stage ends, we compress the KV cache size of each head to a fixed size \( N_{i,1} \). The initial number of heads in each group is consistent. 
% \vspace{-6mm}
\paragraph{Head Information Compression Pattern}
\label{Attention head type screening}

\begin{figure}[ht]
% \vspace{-5mm}
\centering
\includegraphics[width=\columnwidth]{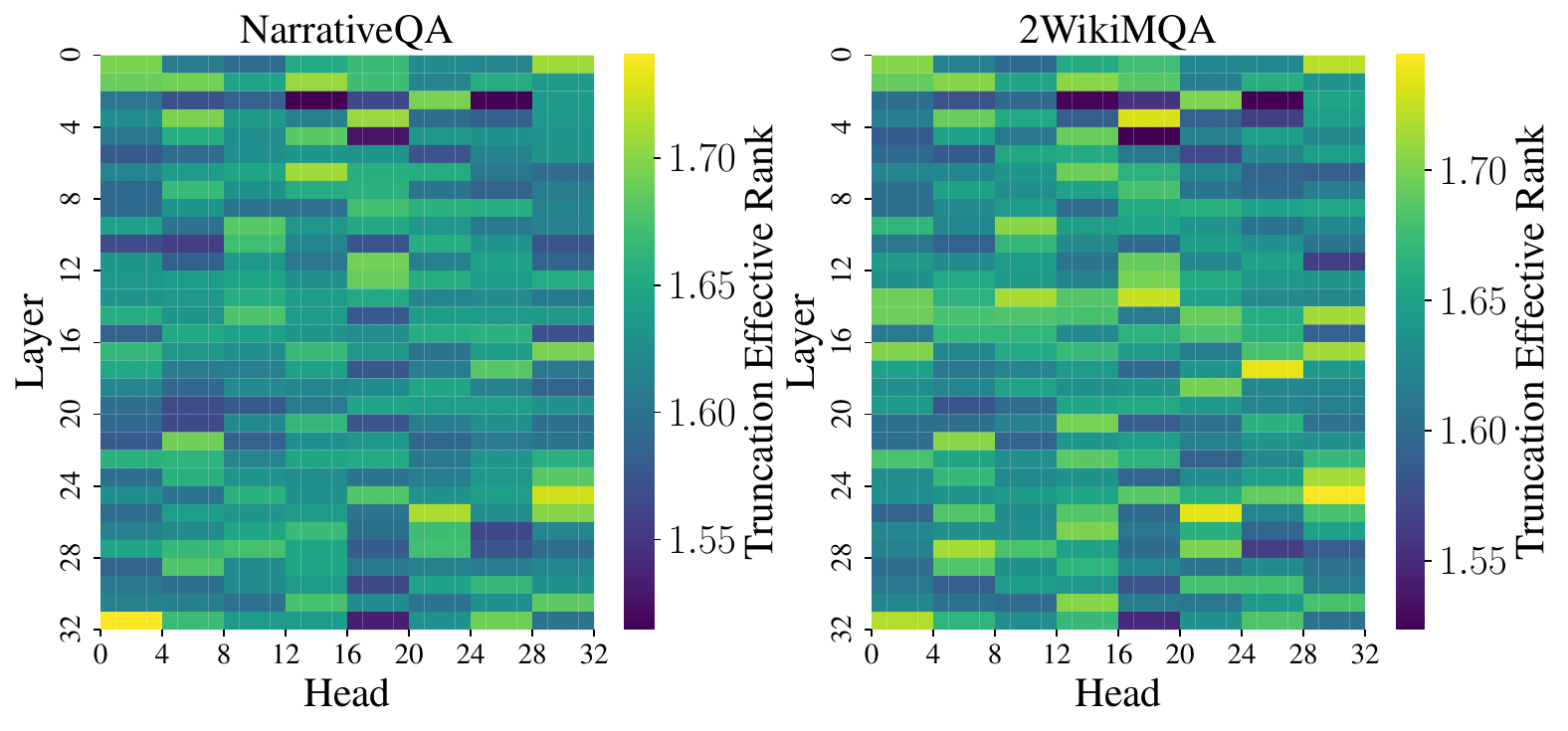}
% \vspace{-8mm}
\caption{The heatmap of \( \text{erank}_{k}(\mathbf{\Sigma}_{Q_m}) \) across different layers and heads in LLaMa3.}
% \vspace{-3mm}
\label{fig:eigenvalue_heatmap_distribution}
\end{figure}

To further reveal the similar information compression patterns within the group of heads, we visualize the \textit{truncated matrix entropy} distribution across different heads of LLaMa3 in Figure~\ref{fig:eigenvalue_heatmap_distribution}. We observe that every four groups of heads exhibit a clearly similar information compression pattern, suggesting they share the same sparsity pattern. This is due to the use of the Group Query Attention (GQA)~\citep{ainslie2023gqa} technique during training, highlighting the significant influence of the training method on the inference approach.

% \vspace{-2mm}
\paragraph{Observation}
 To further explore the mapping between information compression patterns and sparsity patterns, and for the convenience of ablation, we divide the attention heads into two groups: one with a cache size of 96 and the other with a cache size of 32. We set up three configurations: one where the group with a high truncated matrix entropy cache size is set to 96 and the other to 32, one where the group with a low truncated matrix entropy cache size is set to 96 and the other to 32, and one where all attention heads are randomly divided into two groups. These are referred to as High Entropy, Low Entropy, and Random Group, respectively. \textit{As shown in Table~\ref{random_experiment}, we find that heads with a higher truncated matrix entropy require a higher compression ratio (i.e., weaker compression), which often leads to better performance.}

\paragraph{Design}
To leverage this sparsity, heads are ranked by \(\hat{\text{erank}}_{k}(\mathbf{\Sigma}_{Q_m}^{(i,h)})\), grouped into \(M\) (hyperparameter), and then ordered based on the average \(\hat{\text{erank}}_{k}(\mathbf{\Sigma}_{Q_m}^{(i,h)})\) within each group. To calculate the token budget for each head, we define a step size \( \Delta n_g \), progressively reducing the KV cache context size from \( N_{i,1} \) which corresponds to the largest \textit{truncated effective rank}, down to \( N_{i,M} \) as follows:
% \vspace{-2mm}
\begin{equation}
\small
\! N_{i,g+1
} = N_{i,g} - \Delta n_g, \quad g = 1, 2, \dots, M. \!
% \vspace{-1mm}
\end{equation}
where \( N_{i,g} \) represents the context size for the \( g \)-th head group at layer \( i \), and \( \Delta n_g \) is the fixed step size between consecutive groups. During decoding, each head maintains its context window based on the group's token budget \( N_{i,g} \).

% \vspace{-2mm}
\paragraph{Dynamic KV Cache Eviction} We maintain a fixed cumulative attention score window~\citep{li2024snapkv}, \( w_{i,h} \), with a budget size of \( N_{i,g} \), where \( h \) refers to the head. The window \( w_{i,h} \) is computed by summing the attention scores of the last \( l \) tokens from the \( h \)-th head in the \( i \)-th layer, retaining the top \( N_{i,g} \) tokens with the highest scores. As new tokens are generated, the token with the lowest cumulative attention score in \( w_{i,h} \) is evicted.

% \vspace{-2mm}
\paragraph{Retrieval Head} When \(M=2\), we can divide the heads into two groups: retrieval heads and streaming heads. Compared to previous methods~\citep{xiao2024duoattention,tang2024razorattention} that required specially designed datasets and complex searches, we reduce the retrieval head identification from 1.2 hour to 1.6 minute, achieving a 45\(\times\) speedup in retrieval head identification.

% We explore extreme compression rates where heads are removed, excluding them from the forward process. This causes a dimensional mismatch in the attention output due to modified keys and values. To address this, we calculate the cosine distance between attention distributions (over the last \(l\) tokens) of removed and retained heads, then align dimensions by filling the output matrix with the distribution of the most similar retained head.
% \vspace{-2mm}

% \vspace{-1mm}
\section{Experiment}
% \vspace{-1mm}
\subsection{Experimental Settings}
% \paragraph{Setup}
We evaluate three LLMs: Llama2-7B/13B-chat-hf~\citep{touvron2023llama}, Llama-3-8B-Inst~\citep{meta2024llama3}, and Qwen2.5-7B-Instruct~\citep{Yang2024Qwen25TR}. \textsc{UNComp} is compared to KV cache eviction techniques, including H2O~\citep{zhang2024h2o}, PyramidKV~\citep{zhang2024pyramidKV}, SnapKV~\citep{li2024snapkv}, Double-Sparse~\citep{yang2024post}, Quest~\citep{tang2024quest}, StreamLLM~\citep{xiao2023efficient}, RazorAttention~\citep{tang2024razorattention} and CHAI~\citep{agarwal2024chai}. \textsc{UNComp} is assessed on LongBench~\citep{bai2023longbench} and `Needle in a Haystack' task~\citep{liu2024lost}. Results of InfiniteBench~\citep{zhang2024bench} and RULER~\citep{hsieh2024ruler} are provided in Appendix~\ref{Supplementary dataset comparison}.

% \vspace{-1mm}
\subsection{Memory-Constrained Setting}

\paragraph{Main Results} We divide the heads in the KV cache into 8 groups, with \( N_{i,1} = 640\) and \( \Delta n_g \) = 74. For fairness, the KV cache size of all heads in other models is set to 384. From Table~\ref{main-table}, we draw the following conclusions: \textit{i)} \textsc{UNComp} achieves the best performance, especially on LLaMA3, as we reveal in Section~\ref{Attention head type screening} that grouping settings with a compression ratio performs better due to their training with GQA. This leads to certain heads sharing the same sparsity patterns, resulting in a speedup up to 60\% per instance with a 4.68\% compression ratio. \textit{ii)} UNComp exhibits near-lossless performance in certain models, especially compared to the full-size KV cache setting in Llama2-7B/13B-chat-hf, with only a 0.74\% performance loss down to a 9.38\% compression ratio. \textit{iii)} The method most similar to our head grouping approach is \textbf{CHAI}. With a KV cache compression ratio lower than CHAI's 68.55\%, our method achieved 5.4$\times$ faster inference speed.

\begin{table}[ht]
\vspace{-3mm}
\centering
\adjustbox{max width=\columnwidth}{%
\large
\begin{tabular}{c@{\hskip 0.5pt}c@{\hskip 0.5pt}c@{\hskip 0.5pt}c@{\hskip 0.5pt}c@{\hskip 0.5pt}c@{\hskip 0.5pt}c@{\hskip 0.5pt}c@{\hskip 0.5pt}}
\toprule

\multicolumn{8}{c}{\textbf{Llama2-7B-chat-hf, KV Size = 64}} \\

\midrule
\vspace{0.02cm}
\multirow{1}{*}{{\textbf{Methods}}}  
& \multicolumn{1}{c}{\multirow{1}{*}{\textbf{Qasper}}}
& \multicolumn{1}{c}{\multirow{1}{*}{\textbf{Musique}}}
& \multicolumn{1}{c}{\multirow{1}{*}{\textbf{GovReport}}}
& \multicolumn{1}{c}{\multirow{1}{*}{\textbf{TREC}}}
& \multicolumn{1}{c}{\multirow{1}{*}{\textbf{PCount}}}
& \multicolumn{1}{c}{\multirow{1}{*}{\textbf{Lcc}}}
% & \multicolumn{1}{c}{\multirow{1}{*}{\textbf{Average score}}} \\
& \multirow{1}{*}{{\textbf{Average}}} \\
% Llama2-7B-chat-hf
\renewcommand{\arraystretch}{1} % 恢复默认行间距
FullKV & 18.61 & 10.05 & 25.19 & 63.00 & 5.00 & 61.40 & 30.54    \\

\arrayrulecolor[gray]{0.8}
\midrule
\arrayrulecolor{black}
H2O  & 13.84 & 1.33 & 8.57 & 18.00 & 0.50 &  28.86 & 11.85   \\ 
StreamingLLM & 14.26 & 0.79 & 8.37 & 18.50 & 4.00 & 29.81 & 12.62   \\
SnapKV & 15.70 & 6.15 & 11.16 & 40.50 & 5.00 & 43.77 & 20.38 \\ 
PyramidKV & 16.10 & 6.58 & 12.07 & 46.00 & \textbf{5.50} & 46.09 & 22.06  \\ 
Quest & 16.50 & 6.01 & 10.42 & 53.50 & 5.00 & 46.09 & 22.92 \\ 
Double-Sparse & 16.30 & 5.92 & \textbf{16.23} & 51.00 & 5.00 & 42.00 & 22.74 \\
Ours-group-stage & \textbf{17.57} & 6.58 & 15.25 & \textbf{59.00} & 4.50 & 49.75 & \textbf{25.44}  \\ 
Ours-group & 15.40 & \textbf{6.96} & 14.88 & 55.50 & 5.00 & \textbf{50.00} & 24.62  \\ 

\arrayrulecolor[gray]{0.7}
\midrule
\arrayrulecolor{black}
\multicolumn{8}{c}{\textbf{Llama2-7B-chat-hf, KV Cache Size of One Group With Extreme Compression }} \\

\midrule
\vspace{0.02cm}
\multirow{1}{*}{{\textbf{Methods}}}  
& \multicolumn{1}{c}{\multirow{1}{*}{\textbf{Qasper}}}
& \multicolumn{1}{c}{\multirow{1}{*}{\textbf{Musique}}}
& \multicolumn{1}{c}{\multirow{1}{*}{\textbf{GovReport}}}
& \multicolumn{1}{c}{\multirow{1}{*}{\textbf{TREC}}}
& \multicolumn{1}{c}{\multirow{1}{*}{\textbf{PCount}}}
& \multicolumn{1}{c}{\multirow{1}{*}{\textbf{Lcc}}}
% & \multicolumn{1}{c}{\multirow{1}{*}{\textbf{Average score}}}
& \multirow{1}{*}{{\textbf{Average}}} \\
FullKV & 18.61 & 10.05 & 25.19 & 63.00 & 5.00 & 61.40 & 30.54 \\
% Llama2-7B-chat-hf
\arrayrulecolor[gray]{0.8}
\midrule
\arrayrulecolor{black}

Remain-256  & 19.67 & 9.80 & 20.19 & 63.00 & 5.50 & 60.28 & 29.74 \\
Remain-128 & 18.67 & 9.75 & 20.02 & 63.00 & 5.50 & 59.60 & 29.42 \\
Remain-64 & 18.13 & 9.79 & 19.84 & 63.00 & 5.50 & 58.09 & 29.06 \\
Remain-32 & 18.04 & 9.24 & 19.31 & 63.00 & 5.50 & 57.04 & 28.69 \\
Remain-16 & 18.48 & 8.08 & 18.21 & 63.00 & 5.00 & 47.29 & 26.68 \\
Remain-12 & 17.31 & 8.78 & 18.16 & 62.00 & 5.00 & 45.23 & 26.08 \\
Delete-2-heads & 18.30 & 8.58 & 18.98 & 63.00 & 5.50 & 59.38 & 28.96 \\
Delete-4-heads  &13.29 & 7.96 & 19.12 & 62.50 & 5.50 & 53.94 & 27.05 \\
Delete-8-heads  & 12.64 & 6.60 & 9.87 & 63.50 & 3.21 & 37.87 & 22.28 \\
CHAI Delete-2-heads & 18.99 & 10.20 & 23.65 & 69.00 & 4.00 & 55.28 & 30.18 \\
CHAI Delete-4-heads  &16.75 & 9.20 & 20.79 & 57.00 & 4.00 & 50.10 & 26.30 \\
CHAI Delete-8-heads  & - & - & - & - & - & - & - \\
RA Delete-2-heads & 18.27 & 10.84 & 24.87 & 64.00 & 4.00 & 59.87 & 30.31 \\
RA Delete-4-heads & 13.28 & 7.28  & 22.89 & 63.00 & 4.00 & 56.91 & 27.89 \\
RA Delete-8-heads  & 8.91  & 4.29  & 7.19  & 54.00 & 3.00 & 32.98 & 18.40 \\

\bottomrule
\end{tabular}
}
\vspace{-3mm}
\caption{Extreme Compression Ratio. The symbol ‘–’ indicates that the performance is nearly zero.}
\vspace{-6mm}
\label{2_extreme_compression}
\end{table}

% \vspace{-2mm}
\paragraph{Head Pruning}
To further explore the relationship between the information compression pattern and the sparsity pattern, we compare performance under extreme compression settings. For this investigation, we divide the heads into two groups. As shown in Table~\ref{2_extreme_compression}, when the compression ratio of the KV cache is set to 1.56\%, our method shows a substantial enhancement over existing methods.
% \vspace{-1mm}
We further explore the minimum achievable compression ratio for the group with the lower effective rank, as detailed in Table~\ref{2_extreme_compression} where \textit{Ours-remain-tokens-N} indicates the retention of $N$ tokens per attention head, while the other group maintains a KV cache size of 512. Furthermore, \textit{Delete-K-head} denotes the complete pruning of $K$ heads per layer, contingent on the effective rank order. \textit{The results show that our method maintains high accuracy with only 12 tokens retained in streaming heads or after pruning certain heads.} We primarily compare with CHAI, as it also uses head grouping for pruning and maintains strong performance after removing heads. The performance of CHAI crashes when the cache size is only 64 or when 8 heads are removed, so its performance is not provided. This finding further supports the validity of our way of identifying special sparse pattern, such as retrieval heads and streaming heads. We also compared our method with the RazorAttention (RA) approach, which prunes heads based on retrieval heads identification. Although RA shows some advantages under extreme compression rates, its performance still falls short of ours even when 8 heads are pruned.

\begin{table}[ht]
\vspace{-2mm}
    \centering
    \adjustbox{max width=\columnwidth}{%
    \scriptsize
    \begin{tabular}{c c c c c}
    \toprule
    \multirow{2}{*}{\textbf{Methods}} 
    & \multicolumn{4}{c}{\textbf{ NVIDIA A100 80G GPU}} \\
    \cmidrule(lr){2-5} 
    & \textbf{Inference} & \textbf{Prefill} & \textbf{Decoding} & \textbf{Max Memory} \\

    \midrule

    FullKV & 129.13 & 77.34 & 51.79 & 25900 \\
    StreamingLLM & 170.42 & 90.28 & 80.14 & 22908 \\
    H2O & 140.93 & 90.56 & 50.37 & 22936 \\
    PyramidKV & 126.03 & 78.98 & 47.05 & 22938 \\
    SnapKV & 123.71 & 78.71 & 45.00 & 22920 \\
    Quest & 170.59 & 110.33 & 60.26 & 22940 \\
    Ours-group & 137.84 & 79.25 & 58.59 & 22900 \\
    Ours-group-stage & 105.47 & 48.18 & 57.29 & 22988 \\ 
    \bottomrule
    \end{tabular}
    }
    \vspace{-2mm}
    \caption{Time Consumption (s) and Memory Usage (MB) Analysis on an GPU. The open-source code of double-sparse does not implement a true token eviction strategy, so it is not compared here. The initial KV cache size is 384 per layer, with a batch size of 1.}
    \vspace{-5mm}
    \smallskip
    \small
    % \textit{Note:} 
    \label{tab:time_analysis}

\end{table}

% \vspace{-2mm}
\paragraph{Inference Time Latency and Performance}
% \vspace{-1mm}
We analyze the inference time latency and the specific time costs of each component. To achieve reliable time analysis, we synchronize the CPU and GPU clock frequencies to facilitate our measurements. We sample 150 data points from the MultifieldQA and use the Llama2 model to measure the overhead in a single batch on a NVIDIA A100 80G GPU. We analyze the duration of the prefill stage, the decoding duration, the total duration of the stage inference, and the maximum memory usage.

In Table~\ref{tab:time_analysis}, we compare our experimental results, where the prompt and generated token lengths are 3712 and 384, and present the following observations: \textit{i)} The prefill stage takes up more time throughout the inference process. \textit{ii)} UNComp primarily accelerates the prefill stage, achieving up to 60.52\% acceleration over the full-size KV cache in a single batch, mainly benefiting from the compression of \( H_m \) in the first stage. \textit{iii)} For throughput analysis, experiments with a prompt length of 2048 and generation lengths of 8096 for \textit{ultra-long generation} show that FullKV supports a maximum batch size of 6 with a token generation time of 15.67ms. In comparison, UNComp supports a batch size of 32 with a token generation time of 2.45ms, delivering 6.4$\times$ the throughput of FullKV. Details of other prompt lengths and generation text lengths can be found in Appendix~\ref{throughput_analysis_appendix}.

\begin{table}[h] % 'r' 表示右对齐，宽度为 45%
    \vspace{-2mm}
    \centering
    \adjustbox{max width=\textwidth}{%
    \scriptsize
    \begin{tabular}{c@{}c@{}c}
    \toprule
    \multirow{1}{*}{{\textbf{Methods}}}  
    & \multirow{1}{*}{{\textbf{Llama2-4k}}} 
    & \multirow{1}{*}{{\textbf{ Llama3-8k}}} \\
    
    \midrule
    % \multicolumn{18}{c}{\textbf{Llama-2-7B-chat-hf, Extreme-Compressibility and Delete Heads }} \\
    % \arrayrulecolor[gray]{0.8}
    % \midrule
    % \arrayrulecolor{black}
    % \rowcolor{lightgray} 
    FullKV & 98.70 & 84.99   \\
    H2O & 61.14 & 51.56  \\
    StreamingLLM & 50.14 & 42.36  \\
    PyramidKV & 93.24 & 79.08 \\ 
    SnapKV  & 94.50 & 81.27 \\ 
    Quest  & 95.50 & 81.02 \\ 
    Double-Sparse  & 96.30 & 82.12 \\ 
    CHAI  & 97.80  & 84.20 \\ 
    Ours-group & 98.42 & 84.13 \\
    Ours-group-stage  & \textbf{98.80} & 83.73  \\ 
    Ours-group-retrieval-layer  & 98.56 & \textbf{85.02}  \\
    \bottomrule
    \end{tabular}
    }
    \vspace{-2mm}
    \caption{Needle-in-a-haystack results. The interpolation between the retrieval layer and the model's final layer is denoted as Ours-group-retrieval-layer.}
    \vspace{-5mm}
    \label{Needle-in-a-haystack}
\end{table}

% \vspace{-1mm}
\paragraph{Needle in a Haystack Task}
 \textit{The results show that UNComp outperforms FullKV at a 9.38\% compression rate.} Table~\ref{Needle-in-a-haystack} compares Llama2-4k and Llama3-8k, both with a max KV cache size of 384. This indicates that our uncertainty measurement method can identify the special retrieval layers and effectively retrieve key information.

\begin{figure}[ht]
% \vspace{-3mm}
\centering
\begin{subfigure} %[b]
    \centering
    \includegraphics[width=\columnwidth]{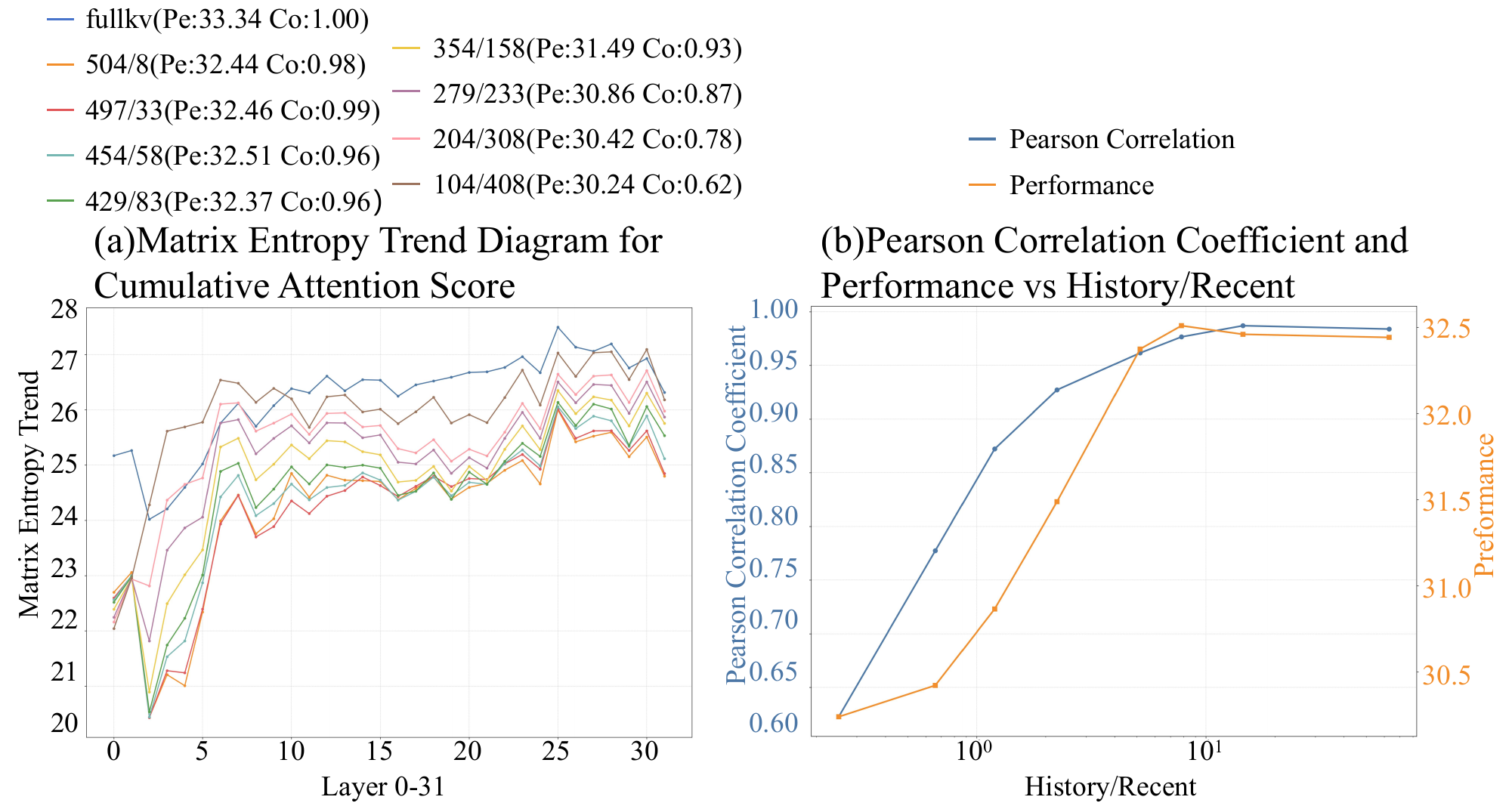}
\end{subfigure}
% \vspace{-8mm}
\caption{Comparison of inter-layer matrix entropy trends for different \(H/R\), where \(H/R\) represents the ratio of the length of historical tokens (\(H\)) to the length of recent tokens (\(R\)), under the fixed cache size setting.}
\label{fig:matrix_entropy_trend}
% \vspace{-5mm}
\end{figure}

\paragraph{The Ratio of Recent Tokens to Historical Tokens} \label{recent_l_token}
\textit{A peculiar phenomenon in KV cache compression: selecting an appropriate proportion of recent tokens can maintain the inter-layer compression trend, thereby enabling optimal performance.} We find that the model's performance is sensitive to the number of recent tokens (the hyperparameter \(l\)), for calculating cumulative attention scores. Our analysis shows that this is due to the special information compression ratio between recent and historical tokens. As shown in Figure~\ref{fig:matrix_entropy_trend}(a), different proportions of historical and recent tokens exhibit distinct trends across layers. The matrix entropy trend of a compressed KV cache, when more similar to the full KV cache across layers, indicates better performance under the same compression ratio. This is validated through experiments using the \textit{pearson correlation coefficient} to quantify the correlation between the trends of different information compression patterns (i.e., changing the proportion of historical tokens and recent tokens across all layers under a fixed token budget). As depicted in Figure~\ref{fig:matrix_entropy_trend} (b), when the inter-layer trend of the compressed KV cache exhibits a higher similarity to the trend of the full-size KV cache, we can achieve optimal performance. 

% \vspace{-2mm}
\section{Conclusion}
% \vspace{-2mm}

We propose \textsc{UNComp}, an uncertainty-aware method for compressing hidden states and KV cache in LLMs. Using \textit{truncated matrix entropy} to measure uncertainty across layers and heads, \textsc{UNComp} adaptively adjusts compression ratios, balancing memory efficiency, computational efficiency, and performance. Experiments show that \textsc{UNComp} achieves up to 60\% speedup in the prefill stage, 6.4$\times$ throughput improvement, and compresses the KV cache down to 4.74\% of its original size, with only a mild performance loss.  \textsc{UNComp} outperforms state of the art in many tasks, demonstrating its effectiveness for LLM inference.

\clearpage
\section*{Limitations}
Our two-stage compression method \textsc{UNComp} demonstrates promising results in accelerating inference and reducing memory overhead. The method performs well on tasks such as the needle-in-a-haystack benchmark, further exploration is needed to assess its effectiveness on tasks involving dense, context-dependent dependencies, such as machine translation or dialogue systems.

\section*{Potential Risks}
While our approach improves the efficiency of long-context inference in LLMs through uncertainty-aware compression, it also introduces new considerations. Dynamically adapting compression based on uncertainty may lead to unintended information loss if uncertainty estimates are miscalibrated, potentially affecting model fidelity in critical tasks. Moreover, the ability to uncover long-range dependencies such as retrieval heads could inadvertently expose internal model mechanisms in ways that may be exploitable. As such, careful validation, transparency in deployment, and alignment with responsible AI practices are essential to ensure these optimizations yield beneficial and trustworthy outcomes without compromising model integrity or user trust.

\section*{Acknowledgements}
This work was supported in part by the Theme-based Research Scheme (TRS) project T45-701/22-R, and in part by the AVNET-HKU Emerging Microelectronics and Ubiquitous Systems (EMUS) Lab.

% \section*{Acknowledgments}

% This document has been adapted
% by Steven Bethard, Ryan Cotterell and Rui Yan
% from the instructions for earlier ACL and NAACL proceedings, including those for
% ACL 2019 by Douwe Kiela and Ivan Vuli\'{c},
% NAACL 2019 by Stephanie Lukin and Alla Roskovskaya,
% ACL 2018 by Shay Cohen, Kevin Gimpel, and Wei Lu,
% NAACL 2018 by Margaret Mitchell and Stephanie Lukin,
% Bib\TeX{} suggestions for (NA)ACL 2017/2018 from Jason Eisner,
% ACL 2017 by Dan Gildea and Min-Yen Kan,
% NAACL 2017 by Margaret Mitchell,
% ACL 2012 by Maggie Li and Michael White,
% ACL 2010 by Jing-Shin Chang and Philipp Koehn,
% ACL 2008 by Johanna D. Moore, Simone Teufel, James Allan, and Sadaoki Furui,
% ACL 2005 by Hwee Tou Ng and Kemal Oflazer,
% ACL 2002 by Eugene Charniak and Dekang Lin,
% and earlier ACL and EACL formats written by several people, including
% John Chen, Henry S. Thompson and Donald Walker.
% Additional elements were taken from the formatting instructions of the \emph{International Joint Conference on Artificial Intelligence} and the \emph{Conference on Computer Vision and Pattern Recognition}.

\nocite{*}
\bibliography{acl_latex}

\clearpage

\appendix
\section*{Appendix}
\section{Implementation details}
\label{Implementation details}

\subsection{Machine Environment}
Main of our experiments are conducted on eight AMD MI210 64G GPUs. Some experiments are conducted on NVIDIA A100 80GB GPUs. Based on our comparison, apart from significant differences in inference speed, the final performance shows slight differences.

\subsection{Model Selection}
In all of our experiments, the model weights are obtained from Hugging Face. Specifically, for the Llama architectures, we utilize the following versions: Llama2-7B employs `meta-llama/Llama-2-7b-chat-hf`, Llama2-13B utilizes `meta-llama/Llama-2-13b-chat-hf`, and Llama3-8B adopts `meta-llama/Meta-Llama-3-8B-Instruct`. For the Qwen architecture, we use the `Qwen/Qwen2.5-7B-Instruct` version.

\subsection{Hyperparameter Setting}
\label{Hyperparameter setting}

We conduct the experiment in a scenario with an average KV cache size of 384 per layer. For the baselines, all hyperparameters are taken from the open-source code repository.

For our method, the experiment is governed by five main hyperparameters, the selection of last \(l\) token’s cumulative attention score,
the threshold  \( \epsilon \), maximum KV cache size of the head in each layer \(N^{'}_{i,1}\),  fixed step size \( \Delta n_g \) and \( \Delta n \) .

For various models, we configure the parameters as follows: \(l\) = 8, \(N^{'}_{i,1}\) = 640, \( \Delta n_g \) = 74 and \( \Delta n \) = 512. Threshold  \( \epsilon \) is set to 0.3 for Llama2-7B, 0.5 for Llama2-13B, 0.4 for Llama3-8B and 0.3 for Qwen2.5-7B.

\section{Details of Evaluation}

Longbench is the first benchmark for assessing the long-context understanding capabilities of large language models in a bilingual and multitask framework. It evaluates multilingual capabilities in both Chinese and English, consisting of six major categories and twenty-one tasks. Key application scenarios include single-document QA, multi-document QA, summarization, few-shot learning, synthetic tasks, and code completion. We use Longbench to evaluate the performance of our method on contextual input tasks. The details of metrics at Table~\ref{evaluation details}.

% \FloatBarrier
% \begin{table*}[h]
% \centering
% \adjustbox{max width=\textwidth}{%
% \scriptsize
% \begin{tabular}{lccc|lccc }
% \toprule
% \multicolumn{1}{c}{{\textbf{Dataset}}}  % Moves the text down
% & \multicolumn{1}{c}{\textbf{Metric}} 
% % & \multicolumn{1}{c}{\textbf{Prefill Stage}}
% & \multicolumn{1}{c}{\textbf{Language}} 
% & \multicolumn{1}{c|}{\textbf{Data Length}}
% & \multicolumn{1}{c}{\textbf{Dataset}}  % Moves the text down
% & \multicolumn{1}{c}{\textbf{Metric}} 
% % & \multicolumn{1}{c}{\textbf{Prefill Stage}}
% & \multicolumn{1}{c}{\textbf{Language}} 
% & \multicolumn{1}{c}{\textbf{Data Length}}
% \\ 

% % Llama-2-7B-chat-hf
% \midrule
% NarrativeQA & F1 & English & 200  & MultiNews & range-l & English & 200 \\
% Qasper & F1 & English &  200 & trec & classification accuracy & English  & 200  \\
% MultifieldQA & F1 & English &  150 & TriviaQA & English & F1 & 200  \\
% HotpotQA & F1 & English &  200 & SAMSum & range-l & English & 200  \\
% 2WikiMQA & F1 & English & 200 & PCount  & exact match accuracy & English & 200  \\
% Musique & F1 & English & 200 & PRe & exact match accuracy & English & 200  \\
% GovReport & range-l & English & 200 & Lcc & edit similarity & Python/C\#/Java & 500  \\
% QMSum & range-l & English & 200 & RB-P  & edit similarity & Python/Java & 500  \\

% \bottomrule
% \end{tabular}
% }

% \caption{The details of statistics in LongBench }
% \label{evaluation details}
% \end{table*}

% \FloatBarrier
\begin{table}[h]
\centering
\adjustbox{max width=\columnwidth}{%
\scriptsize
\begin{tabular}{lccc}
\toprule
\multicolumn{1}{c}{{\textbf{Dataset}}}  % Moves the text down
& \multicolumn{1}{c}{\textbf{Metric}} 
& \multicolumn{1}{c}{\textbf{Language}} 
& \multicolumn{1}{c}{\textbf{Data Length}}
\\

% Llama-2-7B-chat-hf
\midrule
NarrativeQA & F1 & English & 200  \\
Qasper & F1 & English &  200 \\
MultifieldQA & F1 & English &  150 \\
HotpotQA & F1 & English &  200 \\
2WikiMQA & F1 & English & 200 \\
Musique & F1 & English & 200 \\
GovReport & range-l & English & 200 \\
QMSum & range-l & English & 200 \\
MultiNews & range-l & English & 200 \\
trec & classification accuracy & English  & 200  \\
TriviaQA & F1 & English & 200  \\
SAMSum & range-l & English & 200  \\
PCount  & exact match accuracy & English & 200  \\
PRe & exact match accuracy & English & 200  \\
Lcc & edit similarity & Python/C\#/Java & 500  \\
RB-P  & edit similarity & Python/Java & 500  \\

\bottomrule
\end{tabular}
}

\caption{The details of statistics in LongBench }
\label{evaluation details}
\end{table}
\begin{table*}[t]
\centering
\adjustbox{max width=0.86\textwidth}{%
\scriptsize
\begin{tabular}{c c c c c c c}
\toprule
% \midrule
\multicolumn{7}{c}{\textbf{Llama2-7B-chat-hf, KV Cache Size = 384, Prompt+Generate is 3712+384}} \\
\midrule
\textbf{\multirow{2}{*}{Batch Size}} & \multicolumn{2}{c}{\textbf{Ours-group-stage}} & \multicolumn{2}{c}{\textbf{Ours-group}} & \multicolumn{2}{c}{\textbf{FullKV}} \\
\arrayrulecolor[gray]{0.5}
\cmidrule(lr){2-7} 
\arrayrulecolor{black}
 & \textbf{ms/token} & \textbf{max memory used(MB)} & \textbf{ms/token} & \textbf{max memory used(MB)} & \textbf{ms/token} & \textbf{max memory used(MB)} \\
\midrule
1 & 28.055  & 23492  & 28.536  & 23080  & 25.691  & 24690  \\
4 & 7.910   & 37526  & 8.504   & 37516  & 13.806  & 44220  \\
8 & 4.822   & 59014  & 5.436   & 59036  & 11.823  & 72444  \\
10 & 5.340  & 69802  & 5.887  & 69780  & -       & Out-of-Memory \\
12 & 3.994  & 80514  & 4.567   & 80522  & -       & Out-of-Memory \\
\midrule

\multicolumn{7}{c}{\textbf{Llama2-7B-chat-hf, KV Cache Size = 384, Prompt+Generate is 4032+64}} \\
\midrule

\textbf{\multirow{2}{*}{Batch Size}} & \multicolumn{2}{c}{\textbf{Ours-group-stage}} & \multicolumn{2}{c}{\textbf{Ours-group}} & \multicolumn{2}{c}{\textbf{FullKV}} \\
\arrayrulecolor[gray]{0.5}
\cmidrule(lr){2-7} 
\arrayrulecolor{black}
 & \textbf{ms/token} & \textbf{max memory used(MB)} & \textbf{ms/token} & \textbf{max memory used(MB)} & \textbf{ms/token} & \textbf{max memory used(MB)} \\
\midrule
1 & 34.782  & 24298  & 39.231  & 24312  & 36.596  & 24240  \\
4 & 13.671  & 41180  & 18.458  & 41170  & 23.952  & 41146  \\
8 & 10.186  & 66560  & 15.074  & 66580  & 21.944  & 66532  \\
10 & 9.907  & 79198  & 14.603  & 79206  & 21.482  & 79168  \\
12 & 9.464  & 79140  & 14.150  & 79174  & -       & Out-of-Memory \\

\bottomrule
\end{tabular}
}
\caption{Throughput Analysis}
\label{throughput_analysis}
\end{table*}

Additionally, once the data sample is encoded into tokens, if its length exceeds the model's maximum input length, we truncate it by taking equal portions from the beginning and the end.

\section{Throughput Analysis}
\label{throughput_analysis_appendix}

To thoroughly investigate the inference performance of the model, we conduct experiments on an NVIDIA A100 80G GPU. We randomly sample 96 data points from the Wikitext2 dataset, with strict control over the token lengths for both the prompt and generation phases. Detailed analyses of memory usage and throughput are provided in the Table \ref{throughput_analysis} . We can observe that under the setting of a prompt length of 3712 and a generation length of 384 for this short generated text, our method achieves up to 2.96 times throughput.

\section{Ablation Study}
\label{Ablation Study}

\subsection{Truncation Strategy}

\begin{table}[H]  % 'r' 表示右侧, '0.6\textwidth' 表示占据页面宽度的60%
% \vspace{-3mm}
\centering
\adjustbox{max width=1\textwidth}{%
\scriptsize
\begin{tabular}{c@{} c@{} c@{} c@{} c@{} c}
\toprule
\multicolumn{6}{c}{\textbf{Llama-2-7B-chat-hf, KV Cache Size=384}} \\
\midrule
\textbf{Top k} 
& \textbf{ Qasper} 
& \textbf{  QMSum} 
& \textbf{  SAMSum} 
& \textbf{ Lcc} 
& \textbf{ Average} \\
\midrule
Top 16 & 19.28  & 20.38 & 39.45 & 59.72 & 34.71 \\
Top 32 & 19.34  & 20.51 & 39.35 & 59.93 & 34.78 \\
Top 64 & 18.75  & 20.43 & 39.36 & 59.86 & 34.60 \\
Top all & 18.14  & 20.14 & 38.52 & 59.51 & 34.08 \\ 
\bottomrule
\end{tabular}
}
\caption{Truncation Strategy}
\vspace{-2mm}
\label{truncation_strategy}
\end{table}

\begin{table*}[t]
\centering
\adjustbox{max width=\textwidth}{%
\scriptsize
\begin{tabular}{c c c c c c c c}
\toprule

\multicolumn{7}{c}{\textbf{Llama2-7B-chat-hf, KV Cache Size=384}} \\
\midrule

\textbf{Group Number}
& \textbf{KV cache size in different groups}
& \textbf{Qasper} 
& \textbf{HotpotQA} 
& \textbf{QMSum} 
& \textbf{SAMSum} 
& \textbf{Lcc} 
& \textbf{Average} \\

\arrayrulecolor[gray]{0.8}
\midrule
\arrayrulecolor{black}
2 groups & 32/736 & 18.23 & 30.96 & 19.82 & 40.05 & 57.13 & 33.24 \\
3 groups & 32/384/736 & 18.90 & 30.53 & 19.95 & 40.04 & 58.29 & 33.54 \\ 
4 groups & 32/266/502/736 & 19.29 & 30.48 & 20.10 & 41.03 & 59.37 & 34.05 \\
5 groups & 32/208/384/560/736 & 19.58 & 31.17 & 20.72 & 40.61 & 58.93 & 34.20 \\ 
8 groups & 32/132/232/332/436/536/636/736 & 19.34 & 31.04 & 20.16 & 40.92 & 59.48 & 34.19 \\
\arrayrulecolor[gray]{0.8}
\midrule

2 groups & 256/512 & 19.67 & 30.98 & 20.20 & 39.36 & 60.28 & 34.10 \\
3 groups & 256/384/512 & 19.45 & 31.29 & 20.24 & 39.63 & 59.63 & 34.05 \\ 
4 groups & 256/342/427/512 & 19.20 & 30.99 & 20.10 & 39.33 & 59.71 & 33.87 \\
5 groups & 256/320/384/448/512 & 19.71 & 30.95 & 20.22 & 39.60 & 59.99 & 34.09 \\ 
8 groups & 256/296/332/368/404/440/476/512 & 19.55 & 31.02 & 20.59 & 39.10 & 59.39 & 33.93 \\
% \arrayrulecolor[gray]{0.8}
% \midrule
\arrayrulecolor{black}
% 2 groups & 256/512 & 19.67 & 30.98 & 20.20 & 39.36 & 60.28 & 34.10 \\
% 3 groups & 256/384/512 & 19.45 & 31.29 & 20.24 & 39.63 & 59.63 & 34.05 \\ 
% 4 groups & 256/342/427/512 & 19.20 & 30.99 & 20.10 & 39.33 & 59.71 & 33.87 \\
% 5 groups & 256/320/384/448/512 & 19.71 & 30.95 & 20.22 & 39.60 & 59.99 & 34.09 \\ 
% 8 groups & 256/296/332/368/404/440/476/512 & 19.55 & 31.02 & 20.59 & 39.10 & 59.39 & 33.93 \\

\bottomrule
\end{tabular}
}
\caption{Group Number Analysis}
\label{multiple group comparison}
\end{table*}

In this section, we examine truncation strategies, with a focus on evaluating the effectiveness of elbow points. We conduct tests using various elbow points by selecting different top k eigenvalues and compared the results to cases where no elbow points are applied. As demonstrated in Table \ref{truncation_strategy}, the results demonstrate a 0.70\% performance gap between the truncated and untruncated settings, highlighting the efficacy of our approach.

\subsection{Number of Head Groups}

\label{number of groups}

In this section, we analyze the impact of the number of groups on performance. As illustrated in Table \ref{multiple group comparison}, when the KV cache size between groups changes minimally, the maximum performance difference with changes in the number of groups is only 0.23\%. However, when there is a significant disparity between the maximum and minimum KV cache sizes of the groups, increasing the number of groups tends to enhance performance, with a performance improvement of 0.96\%. It indicates that the number of groups is highly correlated with the distribution of KV cache sizes within groups, and the greater the disparity in sparse patterns, the better the performance.

% \FloatBarrier

% \subsection{Compression Ratio Allocation between Head Groups}

% \input{table/different_compression_rates}

% In this section we discuss the allocation of compressibility between groups. Using the same experimental setup as the previous section, we only exchange the KV cache size between the two groups, and find that the text generation exhibits abnormal changes. As demonstrated in Table \ref{different_compression_rates}, the results in the table indicate that text generation exhibited abnormalities in several datasets, with the overall average accuracy decreasing to 5.89\%. This suggests that assigning smaller KV cache sizes to lower entropy groups is effective. Conversely, allocating smaller KV sizes to higher-rank groups leads to significant information loss.

\begin{table*}[ht]
\centering
\adjustbox{max width=\textwidth}{%
\scriptsize
\begin{tabular}{l@{}c@{}c@{}c@{}c@{}c@{}c@{}c@{}c@{}c@{}c@{}c@{}c@{}c@{}c@{}c@{}c@{}c}
\toprule

\multirow{2}{*}{\raisebox{-4ex}{\textbf{Methods}}}  % Moves the text down
& \multicolumn{3}{c}{\textbf{Single-Document QA}} 
& \multicolumn{3}{c}{\textbf{Multi-Document QA}} 
& \multicolumn{3}{c}{\textbf{Summarization}} 
& \multicolumn{3}{c}{\textbf{Few-shot Learning}} 
& \multicolumn{2}{c}{\textbf{Synthetic}} 
& \multicolumn{2}{c}{\textbf{Code}} 
& \multirow{2}{*}{\raisebox{-4ex}{\textbf{Avg.}}} \\

\cmidrule(lr){2-4} \cmidrule(lr){5-7} \cmidrule(lr){8-10} \cmidrule(lr){11-13} \cmidrule(lr){14-15} \cmidrule(lr){16-17}
\setlength{\tabcolsep}{1pt} % 默认值是6pt，减小这个值来减少列间距
& \makebox[1cm]{\raisebox{0.5ex}{\rotatebox{30}{\textbf{NtrvQA}}}} 
& \makebox[1cm]{\raisebox{0.7ex}{\rotatebox{30}{\textbf{Qasper}}}} 
& \makebox[1cm]{\raisebox{0.8ex}{\rotatebox{30}{\textbf{MF-en}}}} 
& \makebox[1cm]{\raisebox{0.4ex}{\rotatebox{30}{\textbf{HotpotQA}}}} 
& \makebox[1cm]{\raisebox{0.3ex}{\rotatebox{30}{\textbf{2WikiMQA}}}} 
& \makebox[1cm]{\raisebox{0.7ex}{\rotatebox{30}{\textbf{Musique}}}} 
& \makebox[1cm]{\raisebox{0.5ex}{\rotatebox{30}{\textbf{GovReport}}}} 
& \makebox[1cm]{\raisebox{0.8ex}{\rotatebox{30}{\textbf{QMSum}}}} 
& \makebox[1cm]{\raisebox{0.6ex}{\rotatebox{30}{\textbf{MultiNews}}}} 
& \makebox[1cm]{\raisebox{0.8ex}{\rotatebox{30}{\textbf{TREC}}}} 
& \makebox[1cm]{\raisebox{0.6ex}{\rotatebox{30}{\textbf{TriviaQA}}}} 
& \makebox[1cm]{\raisebox{0.6ex}{\rotatebox{30}{\textbf{SAMSum}}}} 
& \makebox[1cm]{\raisebox{0.6ex}{\rotatebox{30}{\textbf{PCount}}}}  
& \makebox[1cm]{\raisebox{1.4ex}{\rotatebox{30}{\textbf{PRe}}}}    
& \makebox[1cm]{\raisebox{1.6ex}{\rotatebox{30}{\textbf{Lcc}}}} 
& \makebox[1cm]{\raisebox{1.4ex}{\rotatebox{30}{\textbf{RB-P}}}} \\

% \midrule
% Variance KV (384) & 16.75 & 18.15 & \textbf{32.09} & \textbf{32.42} & 27.29 & 8.50 & 19.46 & \textbf{20.42} & 22.94 & \textbf{62.50} & 84.65 & 38.64 & \textbf{5.50} & \textbf{12.00} & 58.59 & 52.98 & 32.06 \\ 
% Uncomp (384) & \textbf{17.12} & \textbf{19.20} & \textbf{32.09} & 30.99 & \textbf{27.88} & \textbf{9.27} & \textbf{23.20} & 20.10 & \textbf{24.72} & \textbf{62.50} & \textbf{84.72} & \textbf{39.33} & \textbf{5.50} & 11.10 & \textbf{59.71} & \textbf{54.15} & \textbf{32.60} \\ 
% \arrayrulecolor[gray]{0.8}\midrule
% \arrayrulecolor{black}
% Variance KV (64) & 8.75 & 13.58 & 12.24 & 20.27 & 13.38 & 3.89 & 8.76 & 15.73 & 13.98 & 29.50 & 56.22 & 30.35 & \textbf{5.00} & 5.45 & 37.10 & 30.47 & 19.04 \\ 
% Uncomp (64) & \textbf{14.05} & \textbf{15.40} & \textbf{25.56} & \textbf{26.28} & \textbf{21.96} & \textbf{6.96} & \textbf{14.88} & \textbf{18.83} & \textbf{17.58} & \textbf{55.50} & \textbf{81.61} & \textbf{34.74} & \textbf{5.00} & \textbf{5.00} & \textbf{50.00} & \textbf{45.55} & \textbf{27.43} \\

% \bottomrule
% \end{tabular}
% }
% \caption{Comparison of entropy and variance of truncated matrices}
% \label{tab:comparison}
% \end{table*}
\midrule
Variance KV (384) & 16.75 & 18.15 & 32.09 & \textbf{32.42} & 27.29 & 8.50 & 19.46 & 20.42 & 22.94 & 62.50 & \textbf{84.65} & 38.64 & \textbf{5.50} & \textbf{12.00} & 58.59 & 52.98 & 32.06 \\ 
Uncomp (384) & \textbf{17.33} & \textbf{19.34} & \textbf{34.16} & 31.54 & \textbf{28.23} & \textbf{10.04} & \textbf{20.38} & \textbf{20.51} & \textbf{23.33} & \textbf{63.00} & 84.11 & \textbf{39.35} & \textbf{5.50} & 9.50 & \textbf{59.93} & \textbf{54.87} & \textbf{32.57} \\ 
\arrayrulecolor[gray]{0.8}\midrule
\arrayrulecolor{black}
Variance KV (64) & 8.75 & 13.58 & 12.24 & 20.27 & 13.38 & 3.89 & 8.76 & 15.73 & 13.98 & 29.50 & 56.22 & 30.35 & \textbf{5.00} & 5.45 & 37.10 & 30.47 & 19.04 \\ 
Uncomp (64) & \textbf{14.05} & \textbf{15.40} & \textbf{25.56} & \textbf{26.28} & \textbf{21.96} & \textbf{6.96} & \textbf{14.88} & \textbf{18.83} & \textbf{17.58} & \textbf{55.50} & \textbf{81.61} & \textbf{34.74} & \textbf{5.00} & \textbf{5.00} & \textbf{50.00} & \textbf{45.55} & \textbf{27.43} \\

\bottomrule
\end{tabular}
}
\caption{Comparison of entropy and variance of truncated matrices}
\label{tab:comparison}
\end{table*}
\begin{table*}[t]
\centering
\adjustbox{max width=\textwidth}{%
\begin{tabular}{l@{}c@{}c@{}c@{}c@{}@{}c@{}@{}c@{}c@{}@{}c@{}@{}cccccc@{}c}
\toprule
\textbf{RULER(8k)  } & \textbf{niah single 1 } & \textbf{niah single 2 } & \textbf{niah single 3 } & \textbf{niah multikey 1 } & \textbf{niah multikey 2 } & \textbf{niah multikey 3 } & \textbf{niah multivalue } & \textbf{niah multiquery} & \textbf{vt} & \textbf{cwe} & \textbf{fwe } & \textbf{qa 1 } & \textbf{qa 2 } & \textbf{average} \\ 
\midrule
FullKV 8k      & 100.00  & 100.00  & 100.00  & 98.80  & 88.20  & 97.60  & 95.40  & 99.40  & 98.60  & 97.74 & 83.93 & 67.40 & 50.80 & 90.61 \\ 
\arrayrulecolor[gray]{0.8}\midrule
\arrayrulecolor{black}
UNComp         & \textbf{100.00} & \textbf{99.80} & 3.80    & \textbf{99.40} & \textbf{72.80}  & 0.00   & \textbf{81.55} & \textbf{74.75} & 93.88  & 20.78 & 53.93 & 64.40 & 49.60 & \textbf{62.67} \\ 
SnapKV         & \textbf{100.00} & \textbf{99.80} & 1.60    & 98.80  & 72.60  & 0.00   & 78.00  & 71.05  & \textbf{94.36} & 21.16 & 49.60 & \textbf{64.80}& \textbf{50.00} & 61.67 \\ 
PyramidKV      & \textbf{100.00} & 98.40   & 0.00    & 98.40  & 66.00  & 0.00   & 63.60  & 42.55  & 81.96  & 8.16  & 41.00 & 65.00 & 48.60 & 54.90 \\ 
CHAI           & 35.00   & 22.80   & \textbf{23.40}   & 22.00  & 3.80   & 0.60   & 23.40  & 23.80  & 11.24  & 0.66  & 7.00  & 25.80 & 21.80 & 17.02 \\ 
H2O            & 2.80    & 3.80    & 5.80    & 5.40   & 4.00   & \textbf{3.00}  & 4.60   & 5.20   & 4.60   & \textbf{34.60} & \textbf{85.87}& 42.00 & 39.60 & 18.56 \\ 
\midrule
\textbf{RULER(4k)  } & \textbf{niah single 1 } & \textbf{niah single 2 } & \textbf{niah single 3 } & \textbf{niah multikey 1 } & \textbf{niah multikey 2 } & \textbf{niah multikey 3 } & \textbf{niah multivalue } & \textbf{niah multiquery} & \textbf{vt} & \textbf{cwe} & \textbf{fwe } & \textbf{qa 1 } & \textbf{qa 2 } & \textbf{average} \\ 
\midrule
FullKV 4k      & 100.00  & 100.00  & 100.00  & 99.40  & 100.00 & 98.80  & 99.15  & 99.85  & 99.72  & 99.80 & 94.20 & 81.40 & 58.00 & 94.64 \\ 
\arrayrulecolor[gray]{0.8}\midrule
\arrayrulecolor{black}

UNComp         & \textbf{100.00}  & \textbf{99.80}   & 18.80   & 95.60  & \textbf{98.80}  & 0.00   & \textbf{93.00}  & \textbf{93.00}  & \textbf{95.84}  & \textbf{56.06} & \textbf{78.07} & \textbf{81.40} & \textbf{57.20} & \textbf{74.43} \\ 
SnapKV         & \textbf{100.00}  & 99.60   & 8.00    & \textbf{99.40}  & 97.40  & 0.00   & 88.30  & 87.70  & 95.80  & 52.86 & 76.33 & \textbf{81.40} & 56.60 & 72.57 \\ 
PyramidKV       & \textbf{100.00}  & 99.40   & 0.60    & 98.60  & 91.80  & 0.00   & 65.85  & 49.40  & 78.84  & 10.50 & 66.20 & 81.00 & 55.40 & 61.35 \\ 
CHAI           & 44.40   & 54.00   & \textbf{46.60}   & 36.60  & 14.00  & \textbf{7.20}   & 53.40  & 52.60  & 17.16  & 13.00 & 25.60 & 59.40 & 30.20 & 34.94 \\ 
H2O            & 10.40   & 12.60   & 13.00   & 14.60  & 9.20   & 7.00   & 12.25  & 13.15  & 8.64   & 82.94 & 93.00 & 81.80 & 40.00 & 30.66 \\ 
\bottomrule
\end{tabular}%
}
\caption{Performance comparison of methods on RULER benchmark across different context lengths. The first section shows results for an 8k context, while the second section highlights 4k context performance.}
\label{tab:ruler_comparison}
\end{table*}
\begin{table*}[t]
\centering
\adjustbox{max width=\textwidth}{%
\begin{tabular}{lccccccccccccc}
\toprule
\textbf{Method} & \textbf{En.Sum} & \textbf{En.QA} & \textbf{En.MC} & \textbf{En.Dia} & \textbf{Zh.QA} & \textbf{Code.Debug} & \textbf{Code.Run} & \textbf{Math.Calc} & \textbf{Math.Find} & \textbf{Retrieve.PassKey} & \textbf{Retrieve.Number} & \textbf{Retrieve.KV} & \textbf{Average} \\ 
\midrule
FullKV         & 12.55 & 0.27 & 42.79 & 1.00 & 4.04 & 22.34 & 0.00 & 0.00 & 38.57 & 6.27 & 6.44 & 4.80 & 14.38 \\ 
\arrayrulecolor[gray]{0.8}\midrule
\arrayrulecolor{black}
uncomp         & \textbf{11.74} & 0.23 & \textbf{44.98} & 3.80 & 3.00 & 21.57 & 0.00 & 0.00 & \textbf{38.57} & \textbf{6.27} & 6.44 & 0.00 & \textbf{14.77} \\ 
% uncomp\_stage  & 11.91 & 0.26 & 41.92 & 3.88 & 4.00 & 21.57 & 0.00 & 0.00 & 40.00 & 6.78 & 6.61 & 0.00 & \textbf{14.77} \\ 
snapkv         & 11.59 & 0.28 & 42.36 & 1.00 & 4.01 & 21.83 & 0.00 & 0.00 & 38.29 & \textbf{6.27} & 6.61 & 0.00 & 14.22 \\ 
pyramidkv      & 11.34 & 0.23 & 40.61 & 2.50 & \textbf{4.03} & 22.08 & 0.00 & 0.00 & \textbf{38.57} & \textbf{6.27} & \textbf{6.78} & 0.00 & 14.26 \\ 
chai           & 9.69  & \textbf{0.37} & 34.06 & \textbf{8.00 }& 3.26 & \textbf{24.97} & 0.00 & 0.00 & 27.43 & 4.58 & 5.93 & 1.20 & 12.79 \\ 
h2o            & 10.99 & 0.18 & \textbf{44.98} & 3.50 & 3.98 & 22.08 & 0.00 & 0.00 & 37.71 & 1.69 & 1.69 & 0.00 & 14.24 \\ 

\bottomrule
\end{tabular}%
}
\caption{Performance comparison of various methods on InfiniteBench across different tasks, including summarization, QA, mathematical reasoning, and code-related benchmarks. The ``Average" column represents the overall average performance.}
\label{tab:infinitebench}
\end{table*}

\subsection{Matrix Entropy and Attention Variance}

In this section we discuss the grouping policy. We provide the compression ratio estimates based on the variance of attention scores in Table \ref{tab:comparison}, evaluated under two KV cache sizes, 384 and 64. The results clearly highlight our advantages, especially under the budget of 64. This suggests that solely relying on compression ratio estimation based on attention is unreasonable, as attention itself is subject to biases such as the attention sink\citep{xiao2023efficient} and recency bias\citep{peysakhovich2023attention}. It is necessary to introduce additional unbiased compression estimation methods.

\section{Supplementary Dataset Comparison}
\label{Supplementary dataset comparison}

\subsection{RULER}

RULER~\citep{hsieh2024ruler} is a novel synthetic benchmark designed to comprehensively evaluate the capabilities of long-context language models (LMs). Unlike the traditional Needle-in-a-Haystack (NIAH) test, which focuses solely on retrieval tasks, RULER provides flexible configurations to support customized sequence lengths and task complexities. It extends the vanilla NIAH test by introducing diverse variations in the types and quantities of ``needles" and adding new task categories, such as multi-hop tracing and aggregation, to assess capabilities beyond simple context search. The results are shown in Table \ref{tab:ruler_comparison}, where the Llama-3-8B-Instruct model is used, and other settings are consistent with those in the previous section \ref{Hyperparameter setting}. The experiments are conducted on a single A100 80G GPU. Our method demonstrates superior performance, while PyramidKV's relatively poor performance suggests that setting a separate compression rate for each layer may hinder the effective context length of the model. Therefore, an appropriate grouping strategy for the layers is essential.

\subsection{InfiniteBench}

InfiniteBench\citep{zhang2024bench} is a state-of-the-art benchmark designed to evaluate language models' ability to process, understand, and reason over extremely long contexts exceeding 100k tokens. By pushing context lengths 10 times beyond traditional datasets, InfiniteBench aims to advance applications of LLMs and enable high-level interactions in scenarios requiring extensive context comprehension. Results are showed at Table \ref{tab:infinitebench}, where the Llama-3-8B-Instruct model is used, and other settings are consistent with the previous section \ref{Hyperparameter setting}. Our method demonstrates exceptional robustness and is the only approach that surpasses the performance of FullKV.

% \section{Supplementary Method Comparison}

\subsection{Evaluating Generalization on Reasoning Task}
To verify the generalization  of our method, we conducted a comparison of experiments on the GSM8K~\citep{cobbe2021gsm8k} dataset, and the results are shown in the table  \ref{gsm8k}. The experiment demonstrates the superiority of our method in few-shot reasoning performance. We also found that our method significantly outperforms other methods in a zero-shot setting, which suggests that our approach may be able to identify a sparse pattern in the absence of samples, thereby avoiding the loss of reasoning capability.

\begin{table}[H]
\centering
\caption{Based on the input question, half of the tokens are kept at a time, while for few-shot prompts, the 384 KV cache size is consistently maintained.  We compare the experimental results under 6-shot and 12-shot conditions.  The experiments are performed using Llama2-7B-chat-hf.}
\adjustbox{max width=\columnwidth}{
\label{tab:performance_fewshot}
\centering
\begin{tabular}{lccc}
\toprule
\textbf{Method(Number of tokens)} & \textbf{Zero-shot(112))} & \textbf{6-shot(1251)} & \textbf{12-shot(2321)} \\
\midrule
FullKV         & 24.63 & 27.14 & 26.91 \\
StreamingLLM   & 5.68  & 26.46 & 24.68 \\
H2O            & 5.31  & 27.82 & 27.14 \\
CHAI           & 5.69  & 3.87  & 4.06  \\
SnapKV         & 5.53  & 26.61 & 24.48 \\
PyramidKV      & 2.35  & 24.49 & 24.68 \\
Ours-group         & 12.05 & 27.89 & 27.68 \\
Ours-group-stage  & 12.86 & 26.23 & 26.84 \\
\bottomrule
\end{tabular}
}
\label{gsm8k}
\end{table}

% Double-sparse and our method achieved the best performance when using query-based sparse pruning, surpassing our method by 0.08\%. However, when using the key matrix as the criterion for sparse pruning, its performance was worse than ours.

% \section{Pyramidkv with UNComp}

% Based on PyramidKV+Uncomp Group8 heads, which applies PyramidKV using our method of setting different compression ratios for different heads, our method can bring greater improvements to PyramidKV if an appropriate number of groups is chosen. This is because different heads can be categorized as streaming heads and retrieval heads \citep{xiao2024duoattention}. It is reasonable for retrieval heads to compress fewer tokens with their groups. Setting finer-grained groups for compression might harm the performance of retrieval heads. See Table \ref{tab:pyramidkv_uncomp} for the results.
\begin{figure*}[ht]
\centering
\begin{subfigure} %[b]{0.75\textwidth}  % 调整图像大小为50%
    \raggedleft  % 将图像靠右对齐
    \includegraphics[width=\textwidth]{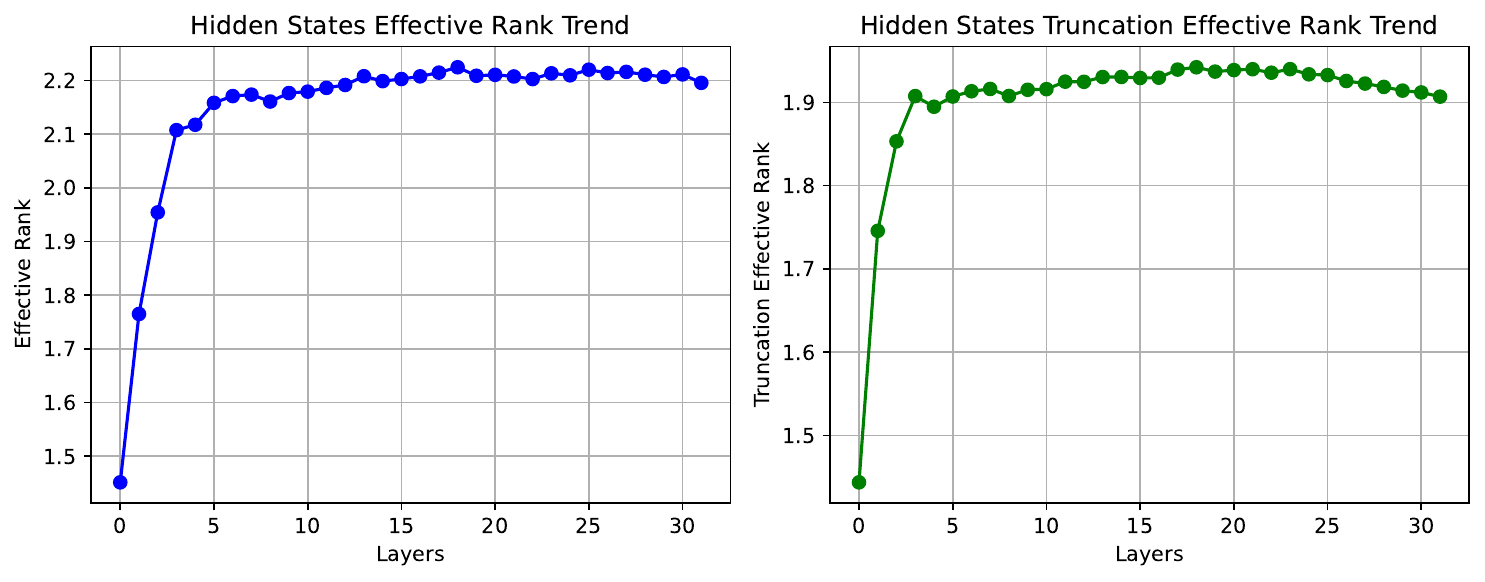}
\end{subfigure}
\caption{Effective rank and truncated effective rank of hidden states across different layers.}
\label{fig:hidden_states_trend}
\end{figure*}

% \vspace{-5mm}
\section{Analysis about Truncated Effective Rank of Hidden States}
\label{hidden states anaylsis}

Figure \ref{fig:hidden_states_trend} shows the entropy trend of sample of the Wikitext2. It can be seen that the matrix entropy of hidden states increases layer by layer, which indicates that the information compression pattern is completely consistent with \(V_m\). This is an interesting phenomenon because the Keys and queries in the KV cache share the same pattern, while the values share the same pattern as the hidden states. We believe this may be due to the positional encodings assigned to both the Queries and Keys. This also reveals a widespread connection between different types of parameters in the model, suggesting that predicting the sparse pattern of one set of parameters using another set is feasible.

% \FloatBarrier

\section{Appendix for Proofs}
\label{Proofs}

The effective rank of \( \mathbf{\Sigma}_{\mathbf{X}} \), denoted $\text{erank}(\mathbf{\Sigma}_{\mathbf{X}})$, is defined as~\citep{roy2007effective}:
\begin{equation}
\small
\label{erank}
\text{erank}(\mathbf{\Sigma}_{\mathbf{X}}) = \exp ( H(\mathbf{\Sigma}_{\mathbf{X}}) ).
\end{equation}

\begin{lemma}
\textit{The rank of the covariance matrix} \(\mathbf{\Sigma}_{\mathbf{X}}\) \textit{is upper bounded by the rank of the input matrix} \(\mathbf{X}\):
%\vspace{-4mm}
\begin{equation}
\small
\label{inequality_erank} 
\text{rank}(\mathbf{\Sigma}_{\mathbf{X}}) \leq \text{rank}(\mathbf{X}).
\end{equation}
\end{lemma}

\begin{lemma}
\textit{Eq.~\eqref{erank} can be interpreted as the dimension of the affine subspace spanned, i.e., the effective dimensionality of the token matrix in the head and layer dimensions. The bounds are:}
%\vspace{-3mm}
% \begin{equation}
% \small
% 1 \leq \text{erank}(\mathbf{\Sigma}_{\mathbf{X}}) \leq \text{rank}(\mathbf{\Sigma}_{\mathbf{X}}) \leq D.
% \end{equation}
\begin{equation}
\small
1 \leq \text{erank}_{\text{trunc}}(\mathbf{\Sigma}_{\mathbf{X}}) \leq \text{erank}(\mathbf{\Sigma}_{\mathbf{X}}) \leq \text{rank}(\mathbf{\Sigma}_{\mathbf{X}}) \leq D.
\end{equation}
\end{lemma}
We use \(\text{erank}(\mathbf{\Sigma}_{\mathbf{X}})\) to quantify the information of token matrix representations, providing a unified uncertainty measure for both the KV cache and hidden states.
% \vspace{-1mm}

% \begin{lemma}
% \textit{Eq.~\eqref{erank} can be interpreted as the dimension of the affine subspace spanned, i.e., the effective dimensionality of the token matrix in the head and layer dimensions. The bounds are:}
% %\vspace{-3mm}
% \begin{equation}
% \small
% 1 \leq \text{erank}(\mathbf{\Sigma}_{\mathbf{X}}) \leq \text{rank}(\mathbf{\Sigma}_{\mathbf{X}}) \leq D.
% \end{equation}
% \end{lemma}

% \vspace{-3mm}
\textbf{Proof Lemma 1} \\
\\
\textbf{Proof.} To derive the von Neumann entropy from the Rényi entropy, we first need to clarify the relationship between the two. The von Neumann entropy can be seen as a special case of the Rényi entropy in the limit where the Rényi parameter $\alpha \to$ 1. The Rényi entropy is defined as:
\begin{equation}
S_\alpha(\mathbf{\Sigma}_{\mathbf{X}}) = \frac{1}{1-\alpha} \log \left( \text{Tr}((\mathbf{\Sigma}_{\mathbf{X}})^\alpha) \right),
\end{equation}
where \(\alpha\) is the order of the Rényi entropy, \(\mathbf{\Sigma}_{\mathbf{X}}\) is the density matrix, and \(\text{Tr}(\rho^\alpha)\) is the trace of the density matrix raised to the power of \(\alpha\). To derive the von Neumann entropy, we need to examine the limit of the Rényi entropy as \(\alpha \to 1\). Let's consider the form of the Rényi entropy:
\begin{equation}
S_\alpha(\mathbf{\Sigma}_{\mathbf{X}}) = \frac{1}{1-\alpha} \log \left( \sum_{i} \sigma_i^\alpha \right),
\end{equation}
where \(\sigma_i\) are the eigenvalues of the density matrix \(\mathbf{\Sigma}_{\mathbf{X}}\). As \(\alpha \to 1\), we can apply L'Hôpital's rule to compute this limit:
\begin{equation}
S(\mathbf{\Sigma}_{\mathbf{X}}) = \lim_{\alpha \to 1} S_\alpha(\rho)
= \lim_{\alpha \to 1} \frac{1}{1-\alpha} \log \left( \sum_i \sigma_i^\alpha \right)
\end{equation}

To proceed, consider the Taylor expansion of \(\sum_i \sigma_i^\alpha\):
\begin{equation}
\begin{split}
\sum_i \sigma_i^\alpha &= \sum_i \sigma_i \cdot e^{(\alpha-1) \log \sigma_i} \\
&\approx \sum_i \sigma_i \left( 1 + (\alpha-1) \log \sigma_i \right) \\
&= 1 + (\alpha-1) \sum_i \sigma_i \log \sigma_i
\end{split}
\end{equation}
Thus,
\begin{equation}
S_\alpha(\mathbf{\Sigma}_{\mathbf{X}}) \approx \frac{1}{1-\alpha} \log \left( 1 + (\alpha-1) \sum_i \sigma_i \log \sigma_i \right)
\end{equation}

As \(\alpha \to 1\), we can use the approximation \(\log(1+x) \approx x\) for small \(x\). Therefore, we get:
\begin{equation}
S_\alpha(\mathbf{\Sigma}_{\mathbf{X}}) \approx -\sum_i \sigma_i \log \sigma_i
\end{equation}
which is exactly the expression for the von Neumann entropy:

\begin{equation}
H(\mathbf{\Sigma}_{\mathbf{X}}) = - \text{Tr}(\mathbf{\Sigma}_{\mathbf{X}} \log(\mathbf{\Sigma}_{\mathbf{X}}))
\end{equation}

\textbf{Proof Lemma 2}\\
% \vspace{1cm}
\\
\textbf{Proof.} In this section, we present a continuous proof of the transformation from the matrix entropy formula to the eigenvalue form.

\begin{equation}
H(\mathbf{\Sigma}_{\mathbf{X}}) = - \text{Tr} \left(  \mathbf{\Sigma}_{\mathbf{X}} \log \left(  \mathbf{\Sigma}_{\mathbf{X}} \right) \right)
\end{equation}

Given that \( \mathbf{\Sigma}_{\mathbf{X}} \) is a symmetric positive definite matrix, we can perform an eigenvalue decomposition:
\begin{equation}
\mathbf{\Sigma}_{\mathbf{X}} = \mathbf{U} \mathbf{\Lambda} \mathbf{U}^\top
\end{equation}
where \( \mathbf{U} \) is an orthogonal matrix composed of eigenvectors, and \( \mathbf{\Lambda} \) is a diagonal matrix whose entries are the eigenvalues \( \sigma_1, \sigma_2, \dots, \sigma_D \). The logarithm of \( \mathbf{\Sigma}_{\mathbf{X}} \) can then be written as:
\begin{equation}
\log(\mathbf{\Sigma}_{\mathbf{X}}) = \mathbf{U} \log(\mathbf{\Lambda}) \mathbf{U}^\top
\end{equation}
where \( \log(\mathbf{\Lambda}) \) is a diagonal matrix whose elements are \( \log(\sigma_1), \log(\sigma_2), \dots, \log(\sigma_D) \). Substituting these into the entropy expression:

\begin{equation}
H(\mathbf{\Sigma}_{\mathbf{X}}) = - \text{Tr} \left( \mathbf{U} \mathbf{\Lambda} \mathbf{U}^\top \mathbf{U} \log(\mathbf{\Lambda}) \mathbf{U}^\top \right)
\end{equation}

Since \( \mathbf{U}^\top \mathbf{U} = \mathbf{I} \), this simplifies to:

\begin{equation}
H(\mathbf{\Sigma}_{\mathbf{X}}) = - \text{Tr} \left( \mathbf{\Lambda} \log(\mathbf{\Lambda}) \right)
\end{equation}

For a diagonal matrix, the trace is the sum of its diagonal elements. Therefore, we have:

\begin{equation}
H(\mathbf{\Sigma}_{\mathbf{X}}) = - \sum_{i=1}^{D} \sigma_i \log(\sigma_i)
\end{equation}

This concludes the proof that the matrix entropy formula can be written as the sum of the eigenvalues of \( \mathbf{\Sigma}_{\mathbf{X}} \).

\textbf{Proof Lemma 3} \\
\\
\textbf{Proof.} Let \(\mathbf{X} \in \mathbb{R}^{n \times p}\) be a matrix representing \(n\) observations and \(p\) variables. The covariance matrix \(\mathbf{\Sigma}_{\mathbf{X}}\) of \(\mathbf{X}\) is defined as:

\begin{equation}
\mathbf{\Sigma}_{\mathbf{X}} = \frac{1}{n-1} \mathbf{X}^\top \mathbf{X}
\end{equation}

The goal is to determine the relationship between the rank of the matrix \(\mathbf{X}\) and the rank of its covariance matrix \(\mathbf{\Sigma}_{\mathbf{X}}\).

The rank of the matrix \(\mathbf{X}\), denoted as \(\text{rank}(\mathbf{X})\), is the number of linearly independent columns in \(\mathbf{X}\), and it satisfies the inequality:

\begin{equation}
\text{rank}(\mathbf{X}) \leq \min(n, p)
\end{equation}

Since the covariance matrix \(\mathbf{\Sigma}_{\mathbf{X}}\) is given by \(\mathbf{\Sigma}_{\mathbf{X}} = \frac{1}{n-1} \mathbf{X}^\top \mathbf{X}\), it is a \(p \times p\) symmetric matrix. The rank of \(\mathbf{\Sigma}_{\mathbf{X}}\), denoted \(\text{rank}(\mathbf{\Sigma}_{\mathbf{X}})\), is determined by the product \(\mathbf{X}^\top \mathbf{X}\). The rank of this product is bounded by the rank of \(\mathbf{X}\), so we have the following inequality:

\begin{equation}
\text{rank}(\mathbf{\Sigma}_{\mathbf{X}}) \leq \text{rank}(\mathbf{X})
\end{equation}

This shows that the rank of the covariance matrix \(\mathbf{\Sigma}_{\mathbf{X}}\) cannot exceed the rank of the original matrix \(\mathbf{X}\). In the case where the number of observations \(n\) is greater than or equal to the number of variables \(p\) (i.e., \(n \geq p\)), and the columns of \(\mathbf{X}\) are linearly independent, the rank of \(\mathbf{X}\) is equal to \(p\), meaning \(\text{rank}(\mathbf{X}) = p\). In this scenario, the matrix \(\mathbf{X}^\top \mathbf{X}\) has full rank, which implies that the covariance matrix \(\mathbf{\Sigma}_{\mathbf{X}}\) will also have full rank. Therefore, we have \(\text{rank}(\mathbf{\Sigma}_{\mathbf{X}}) = p\), and the rank of the covariance matrix is equal to the rank of the original matrix, i.e., \(\text{rank}(\mathbf{\Sigma}_{\mathbf{X}}) = \text{rank}(\mathbf{X})\).

On the other hand, when the number of observations is less than the number of variables (i.e., \(n < p\)), the rank of \(\mathbf{X}\) is constrained by the number of observations, such that \(\text{rank}(\mathbf{X}) \leq n\). Consequently, the rank of the covariance matrix \(\mathbf{\Sigma}_{\mathbf{X}}\) is also limited by \(n\), meaning \(\text{rank}(\mathbf{\Sigma}_{\mathbf{X}}) \leq n\). Since \(n < p\) in this case, the covariance matrix is rank-deficient, and we have \(\text{rank}(\mathbf{\Sigma}_{\mathbf{X}}) < p\).

In general, the rank of the covariance matrix \(\mathbf{\Sigma}_{\mathbf{X}}\) is less than or equal to the rank of the original matrix \(\mathbf{X}\). Specifically, \(\text{rank}(\mathbf{\Sigma}_{\mathbf{X}}) = \text{rank}(\mathbf{X})\) when the number of observations \(n \geq p\) and the columns of \(\mathbf{X}\) are linearly independent. However, when \(n < p\), the covariance matrix \(\mathbf{\Sigma}_{\mathbf{X}}\) will be rank-deficient, such that \(\text{rank}(\mathbf{\Sigma}_{\mathbf{X}}) < p\).

\textbf{Proof Lemma 4}\\
\\
\textbf{Proof.} The entropy $H(\mathbf{\Sigma}_{\mathbf{X}})$ of a set of singular values $\sigma_1, \sigma_2, \dots, \sigma_D$ is given by the formula:
\begin{equation}
H(\sigma_1, \sigma_2, \dots, \sigma_D) = - \sum_{i=1}^{D} \sigma_i \log \sigma_i.
\end{equation}
The trace of \( \mathbf{\Sigma}_{\mathbf{X}} \), \( \text{Tr}(\mathbf{\Sigma}_{\mathbf{X}}) \), is 1. Since entropy measures the uncertainty or disorder in a distribution, we can establish certain bounds for the entropy based on the structure of the singular values. 

First, we note that if the distribution is concentrated entirely at a single value (i.e., all but one of the singular values are zero), then the entropy will be minimized at 0. Specifically:
\begin{equation}
H(1, 0, \dots, 0) = 0.
\end{equation}
On the other hand, the entropy is maximized when the singular values are uniformly distributed. In the case of a uniform distribution over $D$ singular values, we have:
\begin{equation}
\sigma_1 = \sigma_2 = \dots = \sigma_D = \frac{1}{D},
\end{equation}
and the entropy in this case is:
\begin{equation}
H\left(\frac{1}{D}, \frac{1}{D}, \dots, \frac{1}{D}\right) = - D \left( \frac{1}{D} \log \frac{1}{D} \right) = \log D.
\end{equation}
Thus, we have the inequality:
\begin{equation}
0 = H(1, 0, \dots, 0) \leq H(\sigma_1, \sigma_2, \dots, \sigma_D) \leq \log D.
\end{equation}

The \textit{effective rank} is defined as:
\begin{equation}
\text{erank}(\mathbf{\Sigma}_{\mathbf{X}}) = \exp(H(\sigma_1, \sigma_2, \dots, \sigma_D)),
\end{equation}
which quantifies the "effective" number of singular values that are significantly contributing to the rank of the matrix. Since $H(\sigma_1, \sigma_2, \dots, \sigma_D)$ is bounded by $\log D$, it follows that the effective rank is bounded by:
\begin{equation}
1 \leq \text{erank}(\mathbf{\Sigma}_{\mathbf{X}}) \leq D.
\end{equation}
Equality holds at the lower bound if and only if $(\sigma_1, \sigma_2, \dots, \sigma_D) = (1, 0, \dots, 0)$, that is, when all but one singular value is zero. In this case, the singular value vector is:
\begin{equation}
\sigma = \left( \|\sigma\|_1, 0, \dots, 0 \right)^T,
\end{equation}
where $\|\sigma\|_1 = 1$. Hence, $\text{erank}(\mathbf{\Sigma}_{\mathbf{X}}) = 1$.

Next, suppose that only $m$ singular values of $A$ are non-zero for some $m \in \{1, 2, \dots, D\}$. In this case, the rank of $A$ is given by $\text{rank}(A) = m$, and the entropy only depends on the non-zero singular values. Thus, we have:
\begin{equation}
H(\sigma_1, \sigma_2, \dots, \sigma_D) = H(\sigma_1, \sigma_2, \dots, \sigma_m),
\end{equation}
where $\sigma_1, \sigma_2, \dots, \sigma_m$ are the non-zero singular values. Since entropy is maximized when these non-zero singular values are uniformly distributed, we have:
\begin{equation}
H(\sigma_1, \sigma_2, \dots, \sigma_m) \leq \log m.
\end{equation}
Hence, the effective rank satisfies:
\begin{equation}
\text{erank}(\mathbf{\Sigma}_{\mathbf{X}}) \leq m = \text{rank}(\mathbf{\Sigma}_{\mathbf{X}}) \leq D,
\end{equation}
with equality $\text{erank}(\mathbf{\Sigma}_{\mathbf{X}}) = \text{rank}(\mathbf{\Sigma}_{\mathbf{X}})$ if and only if the non-zero singular values are uniformly distributed, i.e., 
% \begin{equation}
% (\sigma_1, \dots, \sigma_m, \sigma_{m+1}, \dots, \sigma_D) = \left( \frac{1}{m}, \dots, \frac{1}{m}, 0, \dots, 0 \right),
% \end{equation}

\begin{equation}
\begin{split}
(\sigma_1, \dots, \sigma_m, \;& \sigma_{m+1}, \dots, \sigma_D) \\
= \;& \left( \tfrac{1}{m}, \dots, \tfrac{1}{m}, \; 0, \dots, 0 \right).
\end{split}
\end{equation}

or equivalently:
\begin{equation}
\sigma = \left( \|\sigma\|_1/m, \dots, \|\sigma\|_1/m, 0, \dots, 0 \right)^T.
\end{equation}
In this case, the effective rank coincides with the actual rank of the matrix, since the singular values contribute equally to the rank.

The truncated entropy $H_{\text{k}}(\mathbf{\Sigma}_{\mathbf{X}})$ of a set of $top$-$k$ singular values $\sigma_1, \sigma_2, \dots, \sigma_k$ is given by the formula:
\begin{equation}
H_{\text{k}}(\sigma_1, \sigma_2, \dots, \sigma_k) = - \sum_{i=1}^{k} \sigma_i \log \sigma_i,
\end{equation}

It is straightforward to obtain that:

\begin{equation}
0 \;\leq\; H_{\text{k}}(\sigma_1, \dots, \sigma_k) \;\leq\; H_(\sigma_1, \dots, \sigma_m),
\end{equation}

Thus:

\begin{equation}
\small
1 \leq \text{erank}_{\text{k}}(\mathbf{\Sigma}_{\mathbf{X}}) \leq \text{erank}(\mathbf{\Sigma}_{\mathbf{X}}) \leq \text{rank}(\mathbf{\Sigma}_{\mathbf{X}}) \leq D.
\end{equation}

Moreover, the truncated entropy is bounded above by the logarithm of the truncation parameter $k$:

\begin{equation}
H_{\text{k}}(\sigma_1, \sigma_2, \dots, \sigma_k) \leq \log k,
\end{equation}

% We now consider the relationship between $k$ and $m$. When $k < m$, we obtain the following inequality:

% {
% \footnotesize
% \begin{equation}
% 1 \leq \text{erank}_{\text{k}} (\mathbf{\Sigma}_{\mathbf{X}}) \leq \text{k} \leq \text{erank}(\mathbf{\Sigma}_{\mathbf{X}}) \leq \text{m} = \text{rank}(\mathbf{\Sigma}_{\mathbf{X}}) \leq D,
% \end{equation}
% }

% When $k \geq m$, the following inequality holds:

% {
% \footnotesize
% \begin{equation}
% \small
% 1 \leq \text{erank}_{\text{k}} (\mathbf{\Sigma}_{\mathbf{X}}) \leq \text{erank}(\mathbf{\Sigma}_{\mathbf{X}}) \leq \text{m} = \text{rank}(\mathbf{\Sigma}_{\mathbf{X}}) \leq \text{k}  \leq D.
% \end{equation}
% }
Then, the following inequality holds:   
\begin{equation}
\left\{
\begin{aligned}
& 1 \leq \text{erank}_{k}(\mathbf{\Sigma}_{\mathbf{X}})
    \leq \min\{k, \text{erank}(\mathbf{\Sigma}_{\mathbf{X}})\} \\
& \quad \leq m = \text{rank}(\mathbf{\Sigma}_{\mathbf{X}}) \leq D,
\quad \text{if } k < m, \\[8pt]
& 1 \leq \text{erank}_{k}(\mathbf{\Sigma}_{\mathbf{X}})
      = \text{erank}(\mathbf{\Sigma}_{\mathbf{X}})\ \\
& \quad \leq m = k = \text{rank}(\mathbf{\Sigma}_{\mathbf{X}}) \leq D,
\quad \text{if } k = m, \\[8pt]
& 1 \leq \text{erank}_{k}(\mathbf{\Sigma}_{\mathbf{X}})
    \leq \text{erank}(\mathbf{\Sigma}_{\mathbf{X}})\ \\
& \quad \leq m = \text{rank}(\mathbf{\Sigma}_{\mathbf{X}}) < k \leq D,
\quad \text{if } k > m.
\end{aligned}
\right.
\end{equation}

\section{A Comprehensive Walk-Through}
\label{algo}
% \vspace{4mm}

\begin{algorithm}[t]
\caption{Head Group Identification (Preparation Phase)}
\label{alg:head-group-preparation}
\begin{algorithmic}[1]
\For{each input sample $x_i$ in the 500-sample batch}
    \State Tokenize $x_i$
    \State Forward $x_i$ through all self-attention layers
    \For{each layer $l$}
        \For{each head $h$}
            \State Compute truncated entropy $E_{l,h}$ (Assign $h$ to one of 8 clusters based on $E_{l,h}$)
        \EndFor
    \EndFor
    \State Save head cluster labels for all layers
\EndFor
\For{each head $h$ across all layers}
    \State Assign final group label by majority vote
\EndFor
\end{algorithmic}
\end{algorithm}

We present the detailed procedure of the algorithm during the preparation phase in Algorithm~\ref{alg:head-group-preparation}. This step is designed to identify consistent attention head behaviors across different layers and input samples by leveraging truncated matrix entropy. Specifically, for each sample in a batch of 500, we tokenize the input and perform a full forward pass through the transformer’s self-attention layers. At each layer, we calculate the truncated entropy \( E_{l,h} \) for every attention head \( h \), which serves as a proxy for information compression.

Each head is then assigned to one of eight clusters according to its entropy value, providing a coarse-grained categorization of its behavior. These per-sample cluster assignments are aggregated across the dataset, and for each head, a final group label is determined by majority voting. This group label serves as a stable representation of the head's functional role and is used in subsequent stages of our method.

\begin{algorithm}[t]
\label{algo1}
\caption{Inference Phase with Dynamic KV Cache Compression}
\label{alg:inference-kv-compression}
\begin{algorithmic}[1]
\State \textbf{Init:} Load head group labels and set the size of kv cache per head

\Statex
\State \textbf{Prefill:}
\For{each transformer layer $l$}
    \If{$l > 0$ and $\text{erank}_{k}(Q_l)$ - $\text{erank}_{k}(Q_{l-1})$  > $\epsilon$} 
        \State Compress hidden states 
    \EndIf
    \For{each head $h$}
        \State Compress the $h$-th head of $\text{kv cache}$ based on group
    \EndFor
\EndFor
\State Forward MLP

\Statex
\State \textbf{Decoding:}
\For{each new token $t$}
    \For{each transformer layer $l$}
        \For{each head $h$}
            \State Concatenate the new token with old the $h$-th head of $\text{kv cache}$ and compute attention using concatenated the $h$-th head $\text{kv cache}$
        \EndFor
    \EndFor
    \State Emit one token and update $\text{kv cache}$
    \State Forward MLP
\EndFor
\end{algorithmic}
\end{algorithm}

In Algorithm~\ref{alg:inference-kv-compression}, we detail the inference phase that incorporates dynamic KV cache compression based on the entropy-driven head groupings derived during the preparation phase. The process begins with initialization, where the group labels for each attention head are loaded, and the cache size is configured accordingly.

During the \textbf{Prefill} stage, for each transformer layer \( l \), the algorithm evaluates the entropy shift between \( Q_l \) and \( Q_{l-1} \). If the difference exceeds a threshold \( \epsilon \), the hidden states are compressed to minimize redundant information. Subsequently, each head's KV cache is selectively compressed based on its assigned group label, balancing efficiency with retention of critical information. The MLP block is then forwarded as usual.

In the \textbf{Decoding} stage, for every new token, each layer performs head-wise attention using a dynamically updated KV cache. The cache is expanded by concatenating the new token with existing head-specific cache entries, enabling efficient autoregressive decoding. After processing through the MLP, a new token is generated, and the KV cache is updated accordingly. This dynamic approach ensures computational and memory efficiency while maintaining model performance.

\end{document}